\documentclass[conference,compsoc]{IEEEtran}
\ifCLASSOPTIONcompsoc

  \usepackage[nocompress]{cite}
\else

  \usepackage{cite}
\fi

\usepackage{graphicx}
\usepackage{multirow}
\usepackage{amsmath}
\usepackage{subfig}
\newcommand{\sys}[1]{{\system{FairDeFace #1}}}

\usepackage{url}
\usepackage{xcolor}

\usepackage{xspace}
\newcommand{\system}[1]{{\small \ensuremath {\mathsf{#1}}}\xspace}
\ifCLASSINFOpdf

\else

\fi                                        

\begin{document}

\title{FairDeFace: Evaluating the Fairness and Adversarial Robustness of \\ Face Obfuscation Methods}

\author{
\IEEEauthorblockN{Seyyed Mohammad Sadegh Moosavi Khorzooghi\IEEEauthorrefmark{2},
Poojitha Thota\IEEEauthorrefmark{2},
Mohit Singhal\IEEEauthorrefmark{3},
Abolfazl Asudeh\IEEEauthorrefmark{4}\\
Gautam Das\IEEEauthorrefmark{2},
Shirin Nilizadeh\IEEEauthorrefmark{2}}
\IEEEauthorblockA{\IEEEauthorrefmark{2} The University of Texas at Arlington 
\IEEEauthorblockA{\IEEEauthorrefmark{3} Northeastern University 
\IEEEauthorblockA{\IEEEauthorrefmark{4} University of Illinois Chicago\\
(seyyedmohammads.moosavikhorzoog,poojitha.thota)@mavs.uta.edu} m.singhal@northeastern.edu, asudeh@uic.edu, gdas@cse.uta.edu, shirin.nilizadeh@uta.edu}
}}

\maketitle

\begin{abstract} 

The lack of a common platform and benchmark datasets for evaluating \emph{face obfuscation} methods has been a challenge, with every method being tested using arbitrary experiments, datasets, and metrics.  
Moreover, while prior work has demonstrated that \emph{face recognition} systems exhibit bias against some demographic groups, there exists a substantial gap in our understanding regarding the fairness of \emph{face obfuscation} methods. 
Providing fair face obfuscation methods can ensure equitable protection across diverse demographic groups, especially since they can be used to preserve the privacy of vulnerable populations, such as activists, survivors of abuse, etc. 
To address these gaps, this paper introduces a comprehensive framework, named \system{FairDeFace}, designed to assess the adversarial robustness and fairness of face obfuscation methods. The framework introduces a set of modules encompassing data benchmarks, face detection and recognition algorithms, adversarial models, utility detection models, and fairness metrics. 
\system{FairDeFace} serves as a versatile platform where any face obfuscation method can be integrated, allowing for rigorous testing and comparison with other state-of-the-art methods. 
In its current implementation, \system{FairDeFace} incorporates 6 attacks, and several privacy, utility and fairness metrics. 
Using \system{FairDeFace}, and by conducting more than 500 experiments, we evaluated and compared the adversarial robustness of seven face obfuscation methods. 
This extensive analysis led to many interesting findings both in terms of the degree of robustness of existing methods and their biases against some gender or racial groups. FairDeFace also uses visualization of focused areas for both obfuscation and verification attacks to show not only which areas are mostly changed in the obfuscation process for some demographics, but also why they failed through focus area comparison of obfuscation and verification.

\end{abstract}


\section{Introduction}
With the proliferation of visual content shared online, there is a growing risk of unauthorized identification and misuse of personal data, particularly through advancements in facial recognition technology. 
The computer security and vision communities have proposed various face obfuscation (also called face de-identification or anonymization) methods to safeguard individuals' privacy and mitigate the risks associated with the widespread dissemination of visual data. 
While traditional methods, such as blurring and pixelation~\cite{hao2020robustness} solely aimed at eliminating identity information to anonymize or obfuscate facial features, the more recent Generative Adversarial Networks (GANs)-based methods like DeepPrivacy~\cite{hukkelaas2019deepprivacy} and CIAGAN~\cite{ciagan} have emerged to balance privacy preservation while maintaining some attributes of the face, such as their emotions and pose.  

The effectiveness of face obfuscation methods has been assessed based on factors like the ability to prevent re-identification, the impact on Face Recognition (FR) algorithms, and the overall usability of the obfuscated face in specific applications, e.g., detecting the emotions of faces in an already anonymized dataset. 
However, the lack of a common platform and benchmark datasets for evaluating face obfuscation methods has been a challenge. Every method tests different threat models, implements arbitrary experiments, and uses arbitrary datasets and metrics, making it difficult to compare different methods objectively and understand their relative strengths and weaknesses. 
For example, the evaluation of CIAGAN's privacy, which focuses on anonymizing faces while maintaining their pose, is conducted using the LFW dataset 
and FaceNet re-identification attacks~\cite{ciagan}. However, this evaluation overlooks other potential attacks, such as verification attacks, its performance is only compared with that of a few other methods~\cite{gafni2019live}. Moreover, its performance in preserving the face pose is not systematically tested.  
This trend is not unique to CIAGAN and is observed across evaluations of various other methods. 

Prior research has extensively studied the fairness of FR systems, showing their strong performance differences across diverse demographic categories, including gender, race, and age~\cite{grother2019face, howard2019effect, klare2012face}. 
Like FR systems, face obfuscation systems might not be fair towards certain demographics, with consequences as severe or even worse than those observed with bias in FR systems. 
For example, obfuscated faces of marginalized or underrepresented groups might be more likely to be reconstructed or identified. 
This can lead to various negative outcomes, including increased risks of privacy infringements, targeted surveillance, or harassment. 

However, to the best of our knowledge, only one study has investigated the fairness properties of a class of face obfuscation systems, i.e., adversarial-based methods~\cite{rosenberg2023fairness}. 
Rosenberg et al.~\cite{rosenberg2023fairness} showed that larger perturbations are needed for the minority groups to have the same obfuscation as the majority groups, which can be unfair because larger perturbations mean less imperceptibility. 
There exists a significant gap in our understanding regarding the fairness of other classes of face obfuscation systems in terms of their performance under various threat models and possible attacks and their ability to preserve utility.   

To address these gaps, this paper introduces \sys{} framework for evaluating the fairness and adversarial robustness of face obfuscation methods. The framework introduces a set of modules encompassing data benchmarks, face detection and recognition algorithms, adversarial models, utility detection models, and fairness metrics. 
Additionally, it incorporates visualization tools using saliency maps and automated feature recognition capabilities through MediaPipe face mesh model to analyze obfuscation patterns. 
The proposed framework serves as a versatile platform, where any face obfuscation method can be integrated for testing and comparison with other state-of-the-art methods. 
Using \sys{}, and by conducting more than 500 experiments, we evaluated and compared the fairness and adversarial robustness of seven face obfuscation methods.

Some of this paper's contributions are:
     (1)~\sys{} is a versatile framework for systematically evaluating and comparing any face obfuscation method with existing ones.
   
    (2)~Created benchmark datasets for evaluating the fairness and robustness of face obfuscation methods. 
 
    (3)~Using \sys{}, we evaluated and compared the performance of seven face obfuscation methods concerning quality, privacy, and utility through state-of-the-art methods. Our analysis resulted in some new findings, e.g., GAN-based methods, especially DeepPrivacy2, having high performance in all the three stages, and low performance of Fawkes even with strong perturbations.
    
    (4) We found that the bias of face obfuscation methods is not necessarily the opposite of the bias in face recognition as Black Males usually had high Bias in both due to large-scale changes done during the obfuscation process.
  
    (5)~Measured how the fairness evaluation is sensitive to the quality of testing datasets, and if the findings are consistent across different datasets. We found that the fairness was more dependent on the obfuscation and face recognition method rather than datasets.  
    
    (6) We proposed and analyzed attribute-preserving rates of GAN-based methods using both open-source and third-party tools.  
    
    (7)~Proposed a novel approach combining saliency maps and automated feature recognition to analyze focus regions of face obfuscation systems and their corresponding face recognition attacks. 
    Our analysis revealed 
    that obfuscation methods, such as Fawkes, rely more on limited or unrelated features than the whole face in robust face recognition systems to change the identity. This is more severe for some demographics, with a lower correlation, which is compatible with the privacy attack results. This can explain their poor performance and bias.
    (8)~Analysing passing rates or successful obfuscated photo generation showed that some obfuscation methods fail for a significant number of photos, especially for Asian Females and Black Males, highlighting the need for more powerful and up-to-date face detection tools such as MTCNN to remove the detection bias part from both obfuscation and recognition.

\section{Related Work}
\label{sec:relwork}

\textbf{Traditional Methods.} 
Prior studies have extensively used pixelation to improve privacy~\cite{boyle2000effects,kitahara2004stealth}, which obfuscates an image by lowering its resolution~\cite{raynal2020image}.
Blurring, another popular method~\cite{besmer2009tagged,ilia2015face,korshunov2013framework}, obfuscates sensitive information by smoothing the face and remove the details using a Gaussian blur~\cite{hill2016effectiveness}. 
These methods, however, are vulnerable to machine learning attacks~\cite{DefeatingImage, hill2016effectiveness} and they do not preserve utility information including age, gender, expression, activity, etc.~\cite{lander2001evaluating}. 
 
\textbf{K-same or K-anonymity Methods.} Prior studies have extensively used the K-same methods~\cite{du2014garp,gross2005integrating,hao2019utility}. 
K-same techniques generate a representative face by averaging all the $k$ faces in a cluster containing faces with similar attributes~\cite{newton2005preserving}. 
However, these de-identified faces suffer in terms of quality and their alignment in the background~\cite{khorzooghi2022stylegan}. 
Additionally, the datasets do not contain enough faces with all attribute combinations, the results might not preserve some of the attributes of the faces~\cite{gross2005integrating,newton2005preserving}. 

\textbf{Differential Privacy.} 
Differential Privacy methods (DP) provide a formulated privacy guarantee~\cite{dwork2006differential},  
such that the attacker is not able to distinguish between the outputs of a differentially-private algorithm. 
 Fan proposed $\epsilon$-differentially private methods in the pixel level and Singular Value Decomposition~\cite{fan2018image,fan2019practical}. 
Todt et. al.~\cite{todt2022fant} used three face obfuscation methods using DP in their experiments: DP-Pixel, DP-Samp~\cite{wang2020videodp} and DP-Snow~\cite{john2020let}. 
However, these methods have a utility-privacy trade-off~\cite{bozkir2023eye}.

\textbf{GAN-based Face Obfuscation.} 
GAN-based methods aim to produce higher-quality obfuscated images while also preserving utility. 
GAN-based methods mainly follow two approaches: (1) conditional inpainting, i.e., cutting out the target face rectangle and constructing a new face by getting help from the target facial attributes, such as pose or identity~\cite{hukkelaas2019deepprivacy,sun2018natural, 8578628,ren2018learning}, and (2) manipulation of facial representation where the facial region removal does not happen~\cite{gafni2019live, wang2021infoscrub, li2019anonymousnet, Wang_2021_CVPR}. 
DeepPrivacy~\cite{hukkelaas2019deepprivacy} and CIAGAN~\cite{ciagan} use conditional GANs for face obfuscation. 
DeepPrivacy2 has improved DeepPrivacy in generating higher quality and  realistic-looking obfuscated images~\cite{hukkelaas2023deepprivacy2}. 

\textbf{Adversarial-based Face Obfuscation Methods.} These methods utilize evasion attack properties~\cite{carlini2017towards, madry2017towards, goodfellow2014explaining}
to add imperceptible perturbations to the face.
FoggySight~\cite{evtimov2020foggysight}, Low-Key~\cite{cherepanova2021lowkey}, Face-Off~\cite{chandrasekaran2021face}, and Fawkes~\cite{shan2020fawkes} are examples of such methods. 
These systems, however, fail to conceal identities from humans. 

\textbf{Obfuscation Methods Evaluation.} 
Previous studies on face obfuscation have usually evaluated their performance on the following three approaches: (1) performing privacy vs. utility analysis~\cite{1640608, 10.1007/11767831_15}, (2) studying human or viewer experience~\cite{hasan2018viewer,li2017effectiveness}, and (3) measuring de-identification robustness against adversarial machine learning attacks~\cite{chattopadhyay2021determining,9320277}.  
However, the main challenge is that there is a lack of benchmark datasets. Prior studies have used different evaluation metrics, e.g., privacy vs. utility or human vs. viewer experience~\cite {1640608,li2017effectiveness}. Additionally, because prior works only evaluate one type of attack~\cite{ciagan}, they often overlook other possible attacks, and hence, it is a challenge when comparing different methods' strengths and weaknesses. 
\section{FairDeFace Framework}
\begin{figure*}[t]
    \centering
    \includegraphics[width=0.8\linewidth]{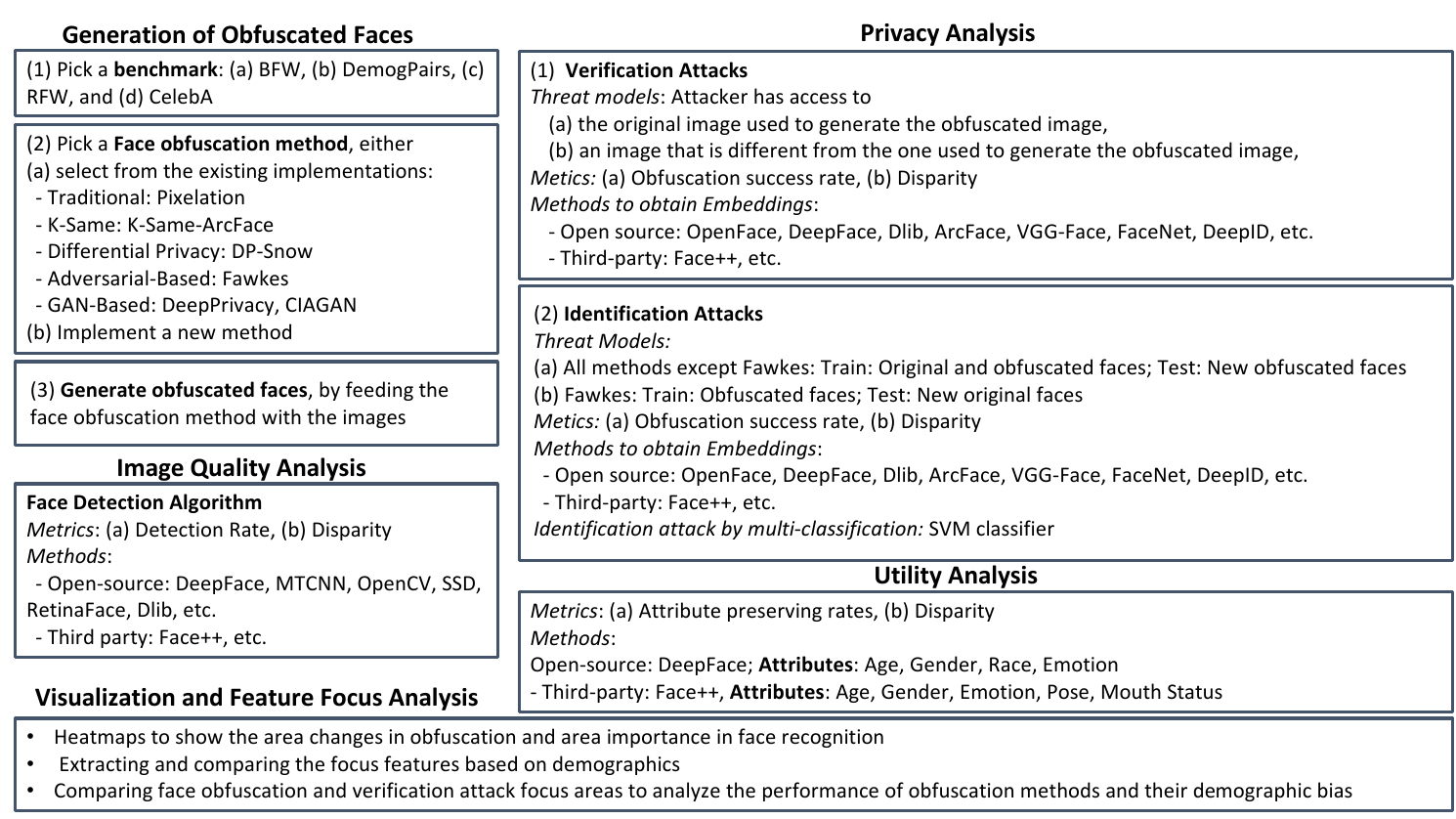}
    \caption{FairDeFace Framework}
    \label{fig:fairdeface}
\end{figure*}
Figure~\ref{fig:fairdeface} illustrates our proposed framework, \sys{}, designed to facilitate the evaluation of adversarial robustness and fairness of face obfuscation methods. 
\sys{} consists of four components: 
(1) \emph{Obfuscated faces generation}, (2) \emph{Image quality analysis}, (3) \emph{Privacy analysis}, (4) \emph{Utility analysis}, and (5) \emph{Visualization and Feature Focus Analysis}. 
Each component comprises several steps, algorithms, and metrics. 
\sys{} is fully implemented and open-sourced for the community to be used, which can be found at this GitHub repository: \url{https://github.com/fairdeface/FairDeface.git.} 
Using our framework, one can pick or implement any face obfuscation method and use our pre-defined commands to test them on any or all of the balanced datasets. 
We ran our experiments on an Intel Xeon W Processor with 184GB of RAM and 4x NVIDIA A5000 GPUs.

\subsection{Generation of Obfuscated Faces} 
The first step in evaluating a face obfuscation method involves taking a set of original facial images and applying the obfuscation technique to generate obfuscated faces. 

\subsubsection{Benchmark Datasets}  
\label{bd}
Several studies have highlighted the link between the quality, diversity, and representativeness of images and the presence of bias in face recognition systems~\cite{grother2019face, howard2019effect,klare2012face}.
Selecting appropriate image datasets is crucial for accurately assessing the effectiveness and fairness of face obfuscation methods. 

\begin{table}[t]
\centering
\caption{Benchmark datasets statistics.} 
\label{table:datasets}
\resizebox{\columnwidth}{!}{%
 \begin{tabular}{llcccccccc} 
\hline
& 
& \multicolumn{2}{c}{White}&\multicolumn{2}{c}{Black}&\multicolumn{2}{c}{Asian}&\multicolumn{2}{c}{Indian}\\
 & & Female & Male & Female & Male & Female & Male & Female & Male\\
     \cline{3-10}
   \multirow{2}{*}{BFW} & \#IDs & 100&100&100&100&100&100&100&100 \\& 
   \#imgs & 2,500&2,500&2,500&2,500&2,500&2,500&2500&2,500 \\
    \multirow{2}{*}{DemogPairs} & \#IDs &100&100&100&100&100&100&0&0  \\&
   \#imgs &1,800&1,800&1,800&1,800&1,800&1,800&0&0 \\
    \multirow{2}{*}{RFW} & \#IDs &\multicolumn{2}{c}{2,959}&\multicolumn{2}{c}{2,995}&\multicolumn{2}{c}{2,492}&\multicolumn{2}{c}{2,984} \\ 
   & \#imgs &\multicolumn{2}{c}{10,196}&\multicolumn{2}{c}{10,415}&\multicolumn{2}{c}{9,688}&\multicolumn{2}{c}{10,308} \\
  \cline{3-10} 
   &&\multicolumn{4}{c}{Male}&\multicolumn{4}{c}{Female}\\

    \multirow{2}{*}{CelebA} & \#IDs &\multicolumn{4}{c}{36,24}&\multicolumn{4}{c}{36,24}  \\ 
   & \#imgs &\multicolumn{4}{c}{36,240}&\multicolumn{4}{c}{36,240}\\
   \hline
\end{tabular}
}
\end{table}
In the literature, we identified three demographic-based datasets: Balanced Faces in the Wild (BFW)~\cite{nmsj-df12-22}, DemogPairs~\cite{8756625}, and Racial Faces in-the-Wild (RFW)~\cite{wang2019racial}.
Among these, \emph{BFW} is the best-balanced dataset, featuring eight demographic groups based on four races (White, Black, Asian, and Indian) and two genders (Male and Female). It includes 2,500 face images per demographic group, representing 100 individuals with 25 images each.
However, we later show that BFW images are of lower quality and may not be an ideal choice for reflecting demographic bias. Their inconsistent results compared to other datasets suggest that image quality has a significant impact on bias.

\emph{DemogPairs} features higher-quality images and includes three races: White, Black, and Asian, but excludes Indian. However, it provides fewer photos per individual (18 compared to 25 in BFW). 
\emph{RFW} is a large race-based dataset with approximately 80,000 images in total. While the number of identities is consistent across demographics, the number of photos per individual varies both within and across demographic groups. 
\emph{CelebFaces Attributes (CelebA)}~\cite{liu2015faceattributes} is a widely used dataset in computer vision, featuring 40 attributes per photo, including gender. To create a balanced dataset by gender, we sampled from CelebA, selecting individuals with at least ten images. This process resulted in 3,624 identities per class, with 10 randomly sampled images for each identity. 
Table~\ref{table:datasets} summarizes the total number of images and identities for each demographic and dataset.

\subsubsection{Face Obfuscation Methods} 
\sys{} includes implementations of seven face obfuscation methods, with at least one method representing each class, as outlined below. Additionally, it enables software and security engineers to implement and test other face obfuscation algorithms. 

\textbf{Pixellation} divides the image into a square grid, and each square is replaced by the average pixel value of the square~\cite{raynal2020image}. In our implementation, we chose the pixelation level as 1/16 for images with sizes of (224*224). 

\textbf{$K$-Same-Pixel-ArcFace}
Newton et al.~\cite{newton2005preserving} developed two methods, $K$-Same-Pixel and $K$-Same-Eigen. In both methods, the obfuscated face is created by averaging $k$ faces: the original face and the top $k-1$ closest faces to it in the PCA domain. The key difference between the two methods lies in how the averaging is performed. 
According to Newton et al.~\cite{newton2005preserving}, any feature vector can be used instead of PCA. For our implementation, we used ArcFace feature vectors from the DeepFace library, as it performed the best among the available face feature extractors.  

\textbf{DP-Snow}~\cite{john2020let} is a differential-privacy-based method in which a percentage of pixels, e.g., $\delta= 0.5$~\cite{todt2022fant}, are replaced with a specific gray value 
, providing differential privacy guarantee of $(0-\delta)$. 
We preferred this method over DP-Pixel and DP-Samp since this method makes the least change in the face having higher utility preserving. 

\textbf{Fawkes} is designed to evade ML recognition by making imperceptible changes to pixel values~\cite{shan2020fawkes}.
As a result, the tracker's system, which has been trained on obfuscated faces, is unable to associate new original faces with the target identity. 
Fawkes is considered a type of poisoning attack, as the tracker only has access to obfuscated, or cloaked, faces for training the system. 
To achieve more robust obfuscation, a higher level of perturbation is required, making the changes less perceptible. Fawkes has three levels of obfuscation: low, medium, and high.

\textbf{DeepPrivacy}~\cite{hukkelaas2019deepprivacy} (DP1) utilizes conditional GANs to ensure privacy is preserved while maintaining the pose and background~\cite{hukkelaas2019deepprivacy}. It consists of a U-NET autoencoder, which is fed the background image (with the face region removed) and incorporates pose information at the bottleneck.

\textbf{DeepPrivacy2} (DP2) is a recent state-of-the-art method that significantly improves upon DeepPrivacy in generating higher-quality, more realistic-looking obfuscated images~\cite{hukkelaas2023deepprivacy2}. Compared to its predecessor, the image resolution has increased from 128x128 to 256x256. 

\textbf{CIAGAN} is a conditional GAN designed to generate anonymized faces while preserving pose and other contextual information, such as the background~\cite{maximov2020ciagan}.
Similar to DP1, CIAGAN employs a U-NET autoencoder, which is provided with background, pose, and identity information. CIAGAN is particularly suitable for video obfuscation, as a desired identity can be input at the bottleneck to generate obfuscation for a specific identity. 

\subsubsection{Obfuscated Faces} 
\begin{figure}[t]
    \centering
    \includegraphics[width=0.7\columnwidth]{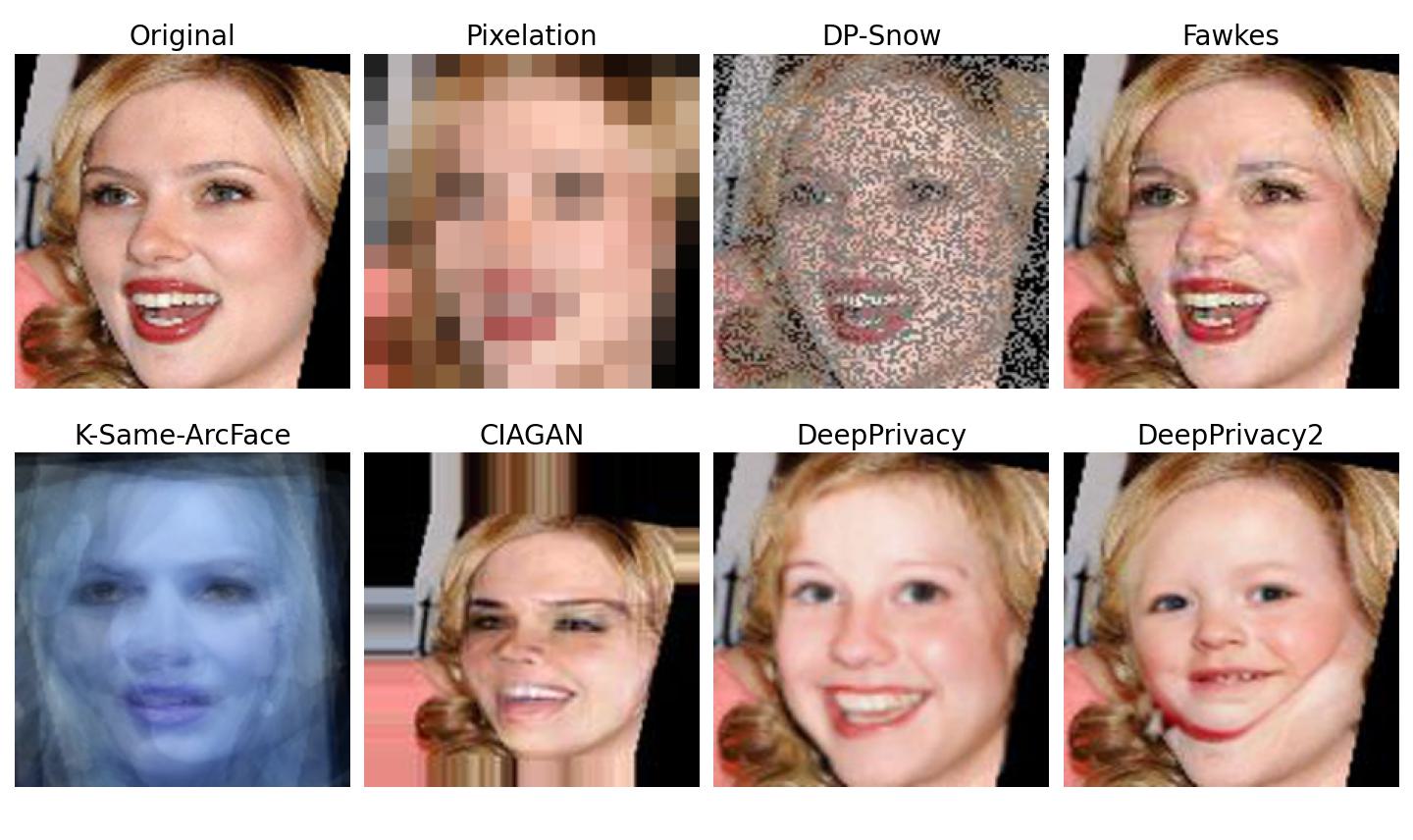}
    \caption{Obfuscated faces for a face example }
    \label{fig:examples-grid}
\end{figure}

We generated obfuscated faces by applying each of the  methods to either all images across the four datasets (for GAN-based methods and DP-Snow) or just BFW and DemogPairs (for the remaining methods). Figure~\ref{fig:examples-grid} presents examples of obfuscated faces for a randomly chosen individual. While a single example may not be fully representative, we observe that some methods produce higher-quality obfuscated faces that better preserve privacy (as perceived by human observers) and retain face attributes.

\textbf{Passing Rate} is a metric evaluates the success of an obfuscation method in generating faces. Some methods may fail to generate obfuscated images for all face images due to errors in face detection, which is a prerequisite for GAN-based and adversarial-based methods. We define the \emph{Passing Rate=$n_o/n$}, where $n_o$ is the number of successfully generated obfuscated faces, and $n$ is the total number of faces. 

\subsection{Fairness Analysis} 
It is crucial for any face obfuscation system to obfuscate faces fairly across all demographics, ensuring that users are treated equitably regardless of their gender or race. 
Fairness in face obfuscation should be evaluated across four key aspects: (1)~success in generating obfuscated faces, (2)~success in producing high-quality images, (3)~effectiveness in ensuring privacy through obfuscation, and (4)~preservation of utility or general facial features. While not all aspects apply to every obfuscation method due to varying objectives, fairness requires that no demographic consistently achieves higher success rates in any aspect.

Ideally, face obfuscation methods should: (1) generate obfuscated faces for all demographics, (2) ensure that the generated obfuscated faces are of high quality across all demographics, (3) ensure equal resistance to re-identification attacks across demographics, and (4) for attribute-preserving methods, maintain corresponding attributes uniformly across all groups. In essence, fairness demands equal passing rates, quality (or detection) rates, obfuscation success rates, and attribute-preserving rates for all demographics. Definitions for the latter two metrics are provided later. 

To the best of our knowledge, one work~\cite{rosenberg2021fairness} has examined the fairness of face obfuscation methods, focusing on a single aspect—privacy— of adversarial-based face obfuscation methods. In contrast, \sys{} provides a comprehensive fairness analysis across multiple types of face obfuscation systems considering all the mentioned aspects. 

\textbf{Sensitive Attributes.} 
\sys{} enables the investigation of fairness across various demographic attributes, including gender, race, and gender-race pairs. Specifically, \sys{} evaluates fairness for \emph{Male} and \emph{Female} categories, and four distinct race groups: \emph{White}, \emph{Black}, \emph{Asian}, and \emph{Indian}. Moreover, it supports privacy and attribute analysis for combinations of gender and race groups, such as Asian Males, Black Females, White Males, and so on.

\textbf{Metrics.} 
There are several competing definitions of fairness~\cite{barocas-hardt-narayanan,narayanan2018translation}. While \sys{} allows for the definition of new fairness metrics, we used \emph{Equality of Opportunity} (EO) to measure bias, which falls under \emph{the group definition of fairness}. According to this definition, the model should provide similar predictions or acceptance rates for various groups, independent of demographic information~\cite{hardt2016equality}. For our analysis, given a set of images from two groups, \emph{a} and \emph{b}, and the four success criteria mentioned above, equality of opportunity is defined as:
\begin{equation}
\label{eq1}
\footnotesize \frac{P(C=1|Y=1, A=a)}{P(C=1|Y=1, A=b)} \leq 1- \epsilon, 
\end{equation}
where $Y=1$ represents the true success, and $C=1$ denotes the success output, which depends on the specific goal of the experiment, i.e., detection success, verification success, identification success, obfuscation success, or attribute-preserving success. \emph{A} represents the sensitive group, which corresponds to a demographic, such as gender, race, or a paired demographic. $\epsilon$ is a slack value, and $P(C=1|Y=1,A=a)$ is the conditional probability, which can be estimated as the average success rate for each group. 
This metric measures the disparity in success rates between the two groups or demographics. For instance, in the case of obfuscation success, it measures whether the probability of \emph{White Females} evading identification by the attacker is similar to that of other race and gender groups. 

\textbf{Threshold:} \textbf{$\epsilon$} is typically set to 0.2~\cite{feldman2015certifying}, but this value can be quite stringent, and prior works have employed different thresholds based on the context~\cite{yee2021image,watkins2024four,bird2020fairlearn}. In \sys{}, we report results for five different $\epsilon$ values: 0.2, 0.15, 0.1, 0.05, and 0.02.
To justify the need for varying thresholds, consider the example of two very different systems: face recognition and stance detection. Face recognition may have high performance, with an accuracy of 0.9, while stance detection may perform poorly, with an accuracy of around 0.5. When evaluating the performance of these systems across two sensitive groups, male and female, we find that the accuracy difference between the groups is 0.1 for both systems. Computing the \emph{Equality of Opportunity} for fairness reveals distinct outcomes. Using a constant threshold of $\epsilon = 0.2$, we find that the well-performing face recognition system shows no bias, while the poorly performing stance detection system exhibits bias.
However, this conclusion may not be entirely accurate, as the face recognition system still performs 10\% worse for one of the groups. Thus, we argue that experts should select an appropriate threshold based on the application, using lower thresholds for systems with higher performance to more accurately capture fairness differences. 

\textbf{Average Bias:}
In addition to using different values of $\epsilon$ to measure bias, \sys{} computes the \emph{Average Bias}~(AB) for a dataset $DS$ with $n$ demographic groups, using a specific $\epsilon_0$ value. The Average Bias is calculated as: $AB(DS|\epsilon_0)=\frac{n_B}{C_2^n}(\%)$, where $C_2^n = \frac{n(n-1)}{2}$ is the total number of demographic group combinations (e.g., Black vs. White), and $n_B$ represents the number of combinations that exhibit bias according to Equation\ref{eq1}.
An average bias of 1 or 100\% indicates that there is bias between every possible combination of demographic groups in the dataset for the given $\epsilon$. This enables us to observe whether different methods show varying levels of bias across different datasets.

\textbf{Demographic Bias:}
Demographic Bias (DB) shows the frequency of the demographic groups in a dataset are biased against a specific demographic for a given $\epsilon$. Given a dataset with $n$ demographics, there will be $n-1$ other demographics to compare their results with.  We define \emph{Demographic Bias} as  $DB(D|\epsilon_0)=\frac{n_D}{(n-1)}$(\%), where $n_D$ is the number of combinations in which other demographics are biased against Demographic $D$, for $\epsilon_0$.

\subsection{Quality Analysis by Face Detection} 
The face detection rate has been frequently used to reflect the quality of obfuscated faces generated by GAN-based obfuscation methods~\cite{hukkelaas2019deepprivacy}. This is because higher quality faces can be detected with higher probability compared to lower quality faces. 
For example, DP1~\cite{hukkelaas2019deepprivacy} used DSFD~\cite{li2019dsfd} face detector for evaluation, showing how the average precision values were almost the same for the original and obfuscated faces, claiming the generated obfuscated faces have the same level of quality as the original faces. 
Similarly, CIAGAN used HOG~\cite{dalal2005histograms} and SSH~\cite{najibi2017ssh} detectors to show attained detection rates after anonymization. 

In addition, face detectors play a crucial role in several obfuscation methods, serving as integral components for pre- or post-processing of images. For instance, they are used to extract the face rectangle for feature selection in methods like Fawkes, or for anonymization in systems such as CIAGAN, DP1, and DP2. Furthermore, face detectors are also essential for privacy and attribute analyses. 

While numerous detection methods are available, their performance varies, and some may introduce biases of their own. Consequently, such biases can be transferred to face obfuscation methods if these detectors are integrated into the process, making it difficult to attribute observed biases solely to the face obfuscation methods, as they may also stem from biases in the face detectors themselves. 
To minimize such biases, we selected the most effective detector across diverse datasets. This evaluation also offers insights into the quality of each dataset: if most or all detectors perform well on a particular dataset, it suggests that the images in that dataset are likely of higher quality. 

\textbf{Metrics:} To investigate fairness across different demographic groups, \sys{} evaluates the Face Detection Rate (FDR) for each demographic. Face detection rate (FDR) is defined as $FDR = \frac{n_D}{n}(\%)$, where $n_D$ represents the number of faces successfully detected out of a total of $n$ faces using a specific detector model. 

\textbf{Methods:} 
\sys{} enables this analysis using both open-source and third-party face recognition systems. It integrates the \emph{DeepFace Library}~\cite{deepface}, a Python library that includes state-of-the-art face detection and recognition models. In our experiments, we evaluated five detection methods: MTCNN~\cite{zhang2016joint}, Dlib~\cite{king2009dlib}, OpenCV~\cite{opencv}, SSD~\cite{liu2016ssd}, and RetinaFace~\cite{deng2019retinaface}. Additionally, \sys{} provides scripts for connecting to the \emph{Face++} API~\cite{faceplusplus}, a third-party system offering services like face detection, comparison, search, and attribute analysis. \sys{} is also designed to support the integration of other detection methods.

\subsection{Privacy Analysis} 
The effectiveness of face obfuscation systems is measured by their robustness against re-identification attacks, such as \emph{verification} and \emph{identification}. These attacks can be implemented under various threat models, depending on the assumptions about the attacker’s knowledge and capabilities. 

\textbf{Verification Attack} is a one-to-one comparison that seeks to determine if two faces, such as an obfuscated and an original face, belong to the same identity.
In \sys{}, verification attacks are implemented using both DeepFace and Face++. DeepFace includes several well-known face recognition methods, such as OpenFace~\cite{baltruvsaitis2016openface}, DeepFace~\cite{taigman2014deepface}, Dlib~\cite{king2009dlib}, ArcFace~\cite{deng2019arcface}, VGG-Face, FaceNet~\cite{schroff2015facenet}, and DeepID~\cite{Sun_2014_CVPR}. 
These systems produce distance scores as output, which are used to compute metrics like Obfuscation Success Rates (OSRs) for obfuscation or True Positive Rates (TPRs) for face recognition. The metrics are calculated using a threshold optimized for the best performance of the face recognition model during its validation phase. 
For Face++, the outputs are confidence scores represented as percentages, where 0\% indicates completely dissimilar faces and 100\% indicates, with high confidence, that the faces belong to the same person. These scores provide an objective measure of the likeness between two images and the probability that they represent the same identity.

\textbf{Identification Attack} involves a one-to-N comparison, where the goal is to identify the obfuscated face by matching it to one of N possible targets. This attack consists of two stages: feature extraction and classification. In \sys{}, identity is represented by feature vectors obtained from any face recognition (FR) method in DeepFace, which can be used for training a classifier. Based on our experiments, (later being explained in Tables~\ref{table:FRs} and~\ref{table:lfw}, we identified \emph{ArcFace} as the most effective FR method in DeepFace. The identity vectors generated by ArcFace were split into training and testing datasets. A fraction of these vectors was used to train a multi-class Support Vector Machine (SVM), while the remaining vectors were reserved for testing the classifier's performance.
\begin{table}[t]
\centering
\caption{\sys{} enables testing face obfuscation methods considering various threat models. $M_4$ and $M_6$ are specific to poisoning-based obfuscation methods.} 
\label{table:assumptions}
\resizebox{\columnwidth}{!}{%
 \begin{tabular}{ll|ccc} 
 \hline
&&\multicolumn{3}{c}{Attacker's Knowledge} \\
Attack Type & Threat Model & Demographic & Original Photos & Same Photo \\  
 \hline 
  \multirow{2}{*}{Verification}& $M_1$  & yes & yes & yes  \\
   &   $M_2$  & yes & yes & no   \\
    \hline 
  \multirow{4}{*}{Identification}&  $M_3$&  no & yes & no  \\
     & $M_4$ &  no & no & no   \\
     & $M_5$&  yes & yes & no  \\
     & $M_6$ &  yes & no & no   \\
  \hline
\end{tabular}
}
\end{table}

\textbf{Threat Models:} The lack of a common platform and benchmark datasets for evaluating face obfuscation methods has been a significant challenge. Each proposed method has been tested under different, often arbitrary, threat models. Particularly in studies involving GAN-based methods, no clear threat model is provided at all, overlooking the need to consider various scenarios from the attacker's perspective. This inconsistency has made it difficult to compare the effectiveness of different obfuscation techniques or assess their robustness under diverse conditions. 
For example, CIAGAN's privacy was evaluated using the race- and gender-imbalanced LFW dataset~\cite{LFWTech} with a single FaceNet re-identification attack, overlooking other attacks like verification and third-party detection~\cite{ciagan}. Similarly, DP1 was tested only on face detection rate against simple methods like pixelation. In contrast, \sys{} enables evaluation of all attack types across diverse, racially and gender-balanced datasets, allowing for a comprehensive comparison of face obfuscation methods under various threat models.

After reviewing existing face obfuscation methods, as it is shown in Table~\ref{table:assumptions}, we identified two threat models ($M_1$ and $M_2$) for verification and four ($M_3$ to $M_6$) for identification attacks. These models differ in the adversary's knowledge and access to obfuscation methods and target photos. For identification, we also considered the attacker's knowledge of the target's demographic~\cite{grother2019face}, with $M_3$ and $M_4$ assuming the attacker is unaware of the demographic, while $M_5$ and $M_6$ assume the attacker knows it.
In verification, the adversary uses the best-performing FR system with an optimal threshold to check if an obfuscated face matches a target. Two scenarios are considered: $M_1$ assumes the attacker has the same face used for obfuscation, and $M_2$ assumes the attacker has other photos of the target, but not the one that is used for obfuscation.

For \emph{identification}, it is assumed that the adversary is aware of the employed face obfuscation method and uses this knowledge to identify the obfuscated face among a set of faces. Specifically, the adversary obfuscates the set of suspected faces with the correct face obfuscation method, extracts their feature vectors using the best-performing FR method, and trains a suitable classifier on the feature vectors. While most obfuscation methods assume the attacker has access to some photos of the target (scenarios $M_3$ and $M_5$), poisoning-based methods such as Fawkes assume a weaker attacker (scenarios $M_4$ and $M_6$). This weaker adversary only has access to the obfuscated (poisoned or cloaked) faces and does not have access to any unmodified faces of the target. 

\textbf{Metrics:} 
Different metrics are used to measure the performance of FRs, including \emph{True Positive Rate} (TPR), also called recall or sensitivity, \emph{True Negative Rate} (TNR), precision, \emph{area under the curve} (AUC), and F1-score.
In face obfuscation, \emph{Obfuscation Success Rate} (OSR)~\cite{rosenberg2023fairness} is used to evaluate performance. It is defined as $OSR = 1 - TPR$, because in face obfuscation, the goal is to evade recognition, which corresponds to achieving high False Negative Rates ($OSR = FNR$).
In other words, to successfully evade verification attacks, the distance score between the obfuscated and the original face should be high, and the confidence score should be low. In the context of verification, OSR is defined as the fraction of positive pairs that the system fails to identify. In identification, OSR is the proportion of times the obfuscated face is not correctly identified, relative to the total number of comparisons. 
Prior work has used OSR and TPRs to evaluate of face obfuscation systems~\cite{ciagan, rosenberg2023fairness}. 
For fairness, Equality of Opportunity (EO) is used to measure the level of bias between any two demographics. This metric is particularly suitable for evaluating bias in face obfuscation, as OSR is related to positive pairs, and Equality of Opportunity focuses specifically on positive pairs as well.

\subsection{Attribute Analysis}
Gross et al.~\cite{gross2006integrating} defined data utility as a set of feature classes derived from applying related utility functions to the face. These features include general facial attributes such as emotion, pose, gender, race, and age, which are not unique and do not directly contribute to identity recognition. In this paper, we focus on specific attributes rather than the broader concept of utility, which is typically balanced with privacy concerns. These attributes are a subset of the overall utility, and their inclusion in the obfuscated face depends primarily on applications. For example, pose and background are commonly preserved features in many face obfuscation systems~\cite{hao2019utility, hukkelaas2019deepprivacy, ciagan, nousi2020deep}. In other applications, features such as lip movements in video obfuscation~\cite{gafni2019live}, as well as emotion~\cite{narula2020preserving, li2021identity, nousi2020deep}, gender~\cite{nousi2020deep} and age~\cite{nousi2020deep} are also maintained. 
Due to the lack of a benchmark and common evaluation functions for assessing how well obfuscation systems preserve utility features, each study has used arbitrary functions and datasets, making it difficult to compare the performance of different methods~\cite{gross2005integrating,chen2021perceptual}. 
Additionally, some works did not evaluate their method's ability to preserve specific attributes, despite claiming claiming to be attribute- or pose-preserving~\cite{hao2019utility, hukkelaas2019deepprivacy, ciagan}.
Moreover, no study has examined the fairness of face obfuscation methods in terms of preserving utility attributes. 
For fair attribute-preserving face obfuscation, it is essential that each attribute be equally preserved across different demographics. 

\textbf{Methods:} \sys{} examines the extent to which each face obfuscation method preserves specific utility attributes. This is done by extracting and comparing the utility features of both the original and obfuscated faces. 
Attribute analysis is typically conducted for GAN-based methods, as they are designed to generate realistic faces. In our experiments, we focus exclusively on testing these methods. \sys{} utilizes facial attribute detectors in DeepFace to estimate attributes such as age, gender, race, and emotion, and in Face++, it estimates features including age, gender, emotion, mouth and eye status, pose, and blurring.

\textbf{Metrics:} 
To determine whether the obfuscated face and the original face share the same utility attributes, we use the \emph{Preserving Rate}, which is calculated in two ways depending on the attribute type: categorical and numerical. For categorical attributes such as gender, race, and emotion, which are represented as distinct categories or states, the Preserving Rate is defined as $PR = \frac{S}{T}$, where $S$ is the number of photo pairs that share the same facial attribute, and $T$ is the total number of photo pairs in a database.
For numerical attributes such as age, which are represented by values between 0 and 100, we define Preserving Rate as: $PR=1-mean(weighted  distances)=1-mean(\frac{\left|Attr_{Obf} -Attr_{Org}\right|}{max(Attr_{Obf}, Attr_{Org})})$, where $Attr_{Obf}$ and $Attr_{Org}$ refer to the attribute values of the obfuscated and original faces in a pair. A lower mean of the weighted distances indicates a higher preserving rate.
The range of the Preserving Rate is [0, 1], where higher values indicate a higher preserving rate for the attribute. The maximum in the denominator is used to accentuate differences for smaller values, as the appearance differences are more pronounced in younger individuals.
Using the preserving rates, we can then calculate EO values to measure fairness across demographic groups.

\subsection{Visualization to Detect the Source of Bias}
After identifying bias in face obfuscation methods, \sys{} analyzes input and output images to uncover the underlying mechanisms that contribute to these disparities. Our approach combines saliency map visualization with automated feature identification to reveal which facial features are most affected during obfuscation and identify bias across demographic groups. 

\begin{figure}[t]
    \centering
    \includegraphics[width=0.9\columnwidth]{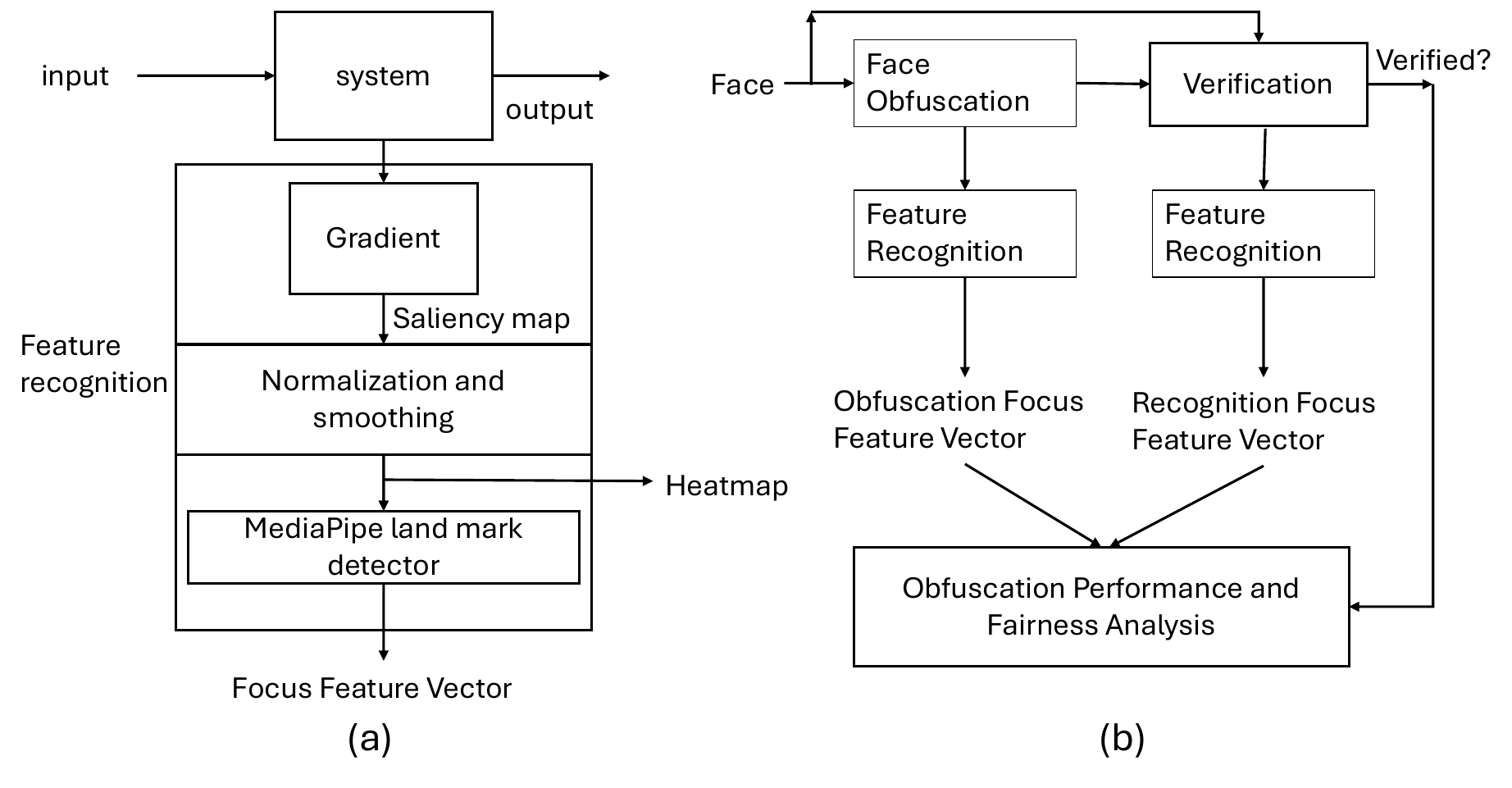}
    \caption{Visualization of Bias Overview }
    \label{fig:visualization}
\end{figure}

Figure~\ref{fig:visualization} shows the flowchart for visualization and feature recognition (a), and obfuscation-verification comparison (b).
\subsubsection{Methodology for Visualization}
\begin{figure}[t]
\centering
\includegraphics[width=0.7\columnwidth]{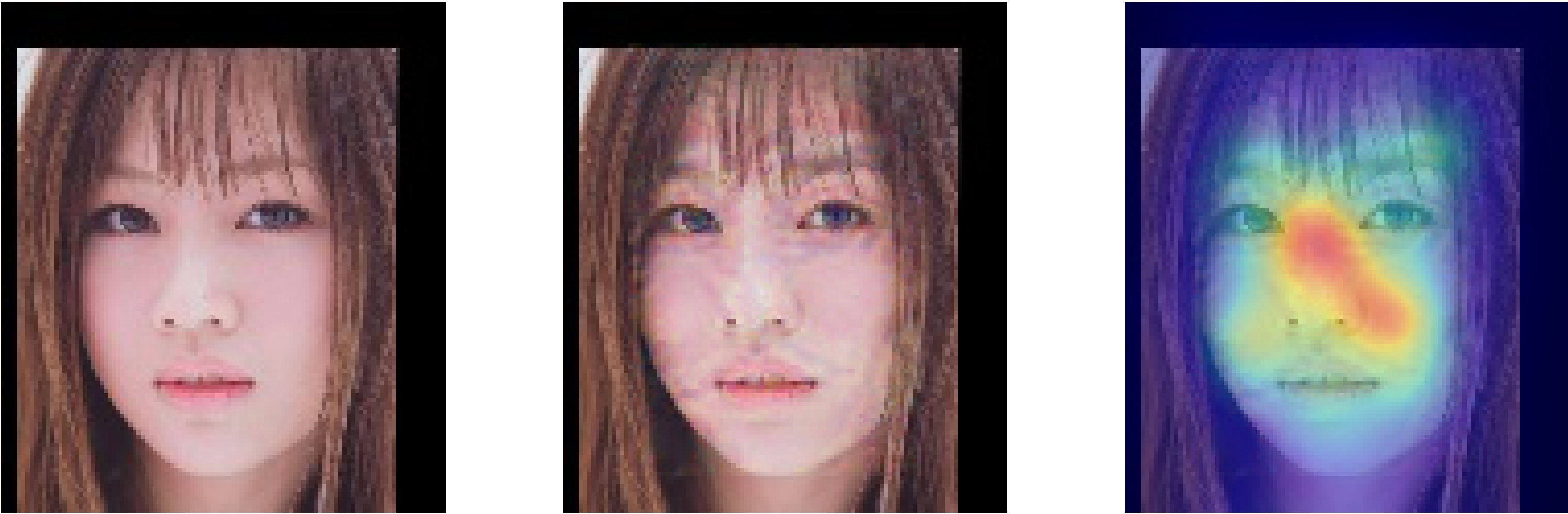}
\caption{Visualization of the Fawkes's Process. Left: Original image. Center: Fawkes-protected image. Right: Saliency map heatmap showing obfuscation focus areas.}
\label{fig:viz_heatmap}
\end{figure}

Our visualization process leverages saliency maps to highlight the areas of the image that are mostly changed during the obfuscation process.
Saliency maps are commonly used in computer vision tasks to identify regions of an image that contribute significantly to a model's decision, as in object recognition and classification~\cite{selvaraju2017grad}. While widely applied in other domains, this technique has never been explored in the context of image obfuscation.
In our approach, a saliency map is generated for each obfuscated photo using a pre-trained feature extractor, which computes the gradients of the model's output with respect to the input image. 
The gradient values are normalized, smoothed, and converted into a heatmap using a jet colormap, where warmer colors indicate high saliency areas. 
The resulting heatmap is overlaid onto the obfuscated image, providing the visualization of where the obfuscation algorithm has focused during obfuscation. 
Figure~\ref{fig:viz_heatmap} illustrates this process, showing an image of Asian Female, its Fawkes-protected version, and the corresponding saliency heatmap, highlighting affected facial areas. 

\subsection{Automated Feature Recognition}
To enable large-scale analysis across different demographic groups and genders, We developed an automated feature recognition process using MediaPipe~\cite{lugaresi2019mediapipe} face mesh model's 468 facial landmarks~\cite{kartynnik2019real} to analyze obfuscation patterns across faces in different demographic groups. 

This innovative approach, was necessary as manually recognizing highly focused features in large datasets is impractical. While existing works~\cite{yang2020fan} often produce heatmaps of important regions, our automated process goes further by mapping these regions to specific facial features, such as nose, forehead, eyes, etc.

The model uses 468 landmarks to capture detailed facial features, with more points allocated to areas of high variability and importance such as the eyes, mouth, and nose. 
We utilize the ``face\_landmarker\_v2\_with\_blendshapes'' task, which detects 468 facial landmarks with high precision~\cite{lugaresi2019mediapipe}. 
Our process maps heatmap regions to specific facial features by applying face landmark detection to visualized areas of high saliency. We quantify obfuscation intensity for each facial region by sampling heatmap values at landmark locations, generating feature vectors for analysis. Results are stored with attributes including race, gender, highest focus feature, and top-5 features with their scores normalized from 0 to 1.

\subsection{Focus Feature correlation}
To fully understand the effectiveness of face obfuscation techniques, we compared the focus areas of the obfuscation systems with their corresponding verification attack. 
To recognize and quantify the features covered during the face recognition process at a large scale, we apply similar methods described in the previous subsection and obtain a list of features. 
We calculated Pearson correlation coefficients between feature vectors extracted from their respective saliency maps to quantify the differences between obfuscation and recognition focus areas. For each image, we aggregated gradient values from the saliency maps across all facial regions detected by MediaPipe landmarks, creating feature vectors representing each model's focus patterns. These vectors were then compared using Pearson correlation, which provides a measure (-1 to 1) of how similarly the two models prioritize facial features. A correlation close to 0 indicates the models focus on different facial regions, while values closer to 1 suggest similar focus patterns.

\section{Results}

\label{resss}
We utilized \sys{} to evaluate the effectiveness and fairness of seven face obfuscation methods. Over 500 experiments were conducted to perform detection, verification, and identification attacks, as well as attribute analysis on these systems. Additionally, we investigated sources of bias in two of these methods. While numerous figures were generated to illustrate the results, they could not all be included in the paper due to space constraints. Therefore, we present only the most noteworthy findings in the paper. All figures, results, and the framework's source code are available in our anonymous GitHub repository:
\url{https://github.com/fairdeface/FairDeface.git}. 
Table~\ref{table:overall} presents the average results for verification, identification, and utility analyses across all datasets, comparing various obfuscation systems with the original, unobfuscated images. These results are  explained in the following sections.

\textbf{}

\begin{table}
\centering
\caption{Overall results of the FairDeface analysis in \% across different datasets} 
\resizebox{\linewidth}{!}{%
 \begin{tabular}{l|c c c cc} 
  \hline
 \textbf{Algorithms}& \textbf{Passing Rate}&\textbf{Detection}&\textbf{Veri.}&\textbf{Iden. (S1)}&\textbf{Iden. (S2)}\\
     \hline
   Unobfuscated &N/A&100&83.4&83&85 \\
   Pixelation&100&46&93&99&96 \\
   DP-Snow&100&90&58&79&73 \\
   Fawkes-High&72&100&2&37&31\\
   K-Same&99&89&42&94&93 \\
   CIAGAN&85&100&39.4&91&97 \\
   DP1&100&96&69.5&97&94\\
   DP2&100&100&76&96&94\\
    \hline
\end{tabular}
\label{table:overall} 
}
\end{table} 

\subsection{Obfuscated Datasets} 
We applied GAN-based methods to all four datasets, obtaining results for verification and attribute analysis across the datasets. However, identification attacks were conducted only on BFW and DemogPairs due to the large number of identities in the other two datasets, which would have resulted in extended training times. 
The four non-GAN-based methods were applied only to BFW and DemogPairs for the sake of brevity. 

\textbf{Passing Rates:} 
 
Table~\ref{table:overall} shows that systems employing a face detector may achieve a PR lower than 100\%. For instance, the K-Same method used MTCNN, which resulted in a PR of 99\%.
In contrast, DP1 and DP2 successfully generated obfuscated faces for all photos, achieving 100\% PR. However, CIAGAN and Fawkes demonstrated lower passing rates of 85\% and 72\%, respectively, due to the limitations of their face detectors, Dlib and MTCNN. 

\emph{Fawkes} failed on 16\% of BFW photos and 51\% of DemogPairs photos.

\textbf{Summary:} Among the methods that employed a face detector, \emph{Fawkes} and \emph{CIAGAN} exhibited significantly lower passing rates. Both methods frequently demonstrated unfairness towards \emph{Asian Females} and \emph{Black Males}.

\subsection{Detection}
Face detectors are an integral component of our privacy and utility evaluation. By identifying faces in an image, they return bounding box coordinates that localize the detected faces. 
We evaluated face detectors provided in the DeepFace library to identify the most suitable one for our privacy and utility assessments. 
We tested five face detectors—OpenCV, Dlib, SSD, RetinaFace, and MTCNN—on the original datasets, analyzing their performance and potential biases. Table~\ref{table:detection_datasets-ratio} in Appendix shows the results. In the following, we discuss the most noteworthy results. 

 \subsubsection{Demographic Bias} 
\textbf{Detectors:} Among the evaluated detectors, \emph{MTCNN} performed the best, achieving nearly 100\% detection rates across all datasets, followed by Dlib and OpenCV. Its consistent detection rates suggest minimal bias, which is crucial for our purposes. Since we use this tool to evaluate the fairness of face obfuscation methods, the absence of detectable bias ensures that any observed bias in the obfuscation methods cannot be attributed to the detector itself. 
\emph{OpenCV} consistently showed bias against Black individuals and males, with Black males being the most affected. Asians were the second-most affected demographic in the corresponding datasets. \emph{Dlib} was frequently unfair for $\epsilon \leq 0.1$, particularly against males, with the bias most pronounced for Black and White males in BFW and DemogPairs, and for Africans in RFW. \emph{SSD} and \emph{RetinaFace} exhibited no bias in any dataset, except for BFW, which is likely more related to the dataset itself than the detector.

\textbf{Datasets:} CelebA showed zero bias for all the methods and $\epsilon$ except for OpenCV at $\epsilon=0.02$ for the original datasets and DP1. This could be attributed to the higher image quality, the presence of only two demographics.
Interestingly, \emph{BFW} showed the lowest detection rates across all detectors, suggesting that despite its more balanced representation and broader demographics, the quality of its images may not be high. Notably, SSD and RetinaFace emphasized this issue more prominently, showing extremely low detection rates for BFW—nearly zero for SSD and only 28\% for RetinaFace—compared to almost 100\% detection rates for the other datasets. 

\textbf{Obfuscation Methods (Quality Analysis):}
We employed \emph{MTCNN} on obfuscated images generated by each obfuscation method to measure the percentage of images containing at least one detected face. This metric serves as an indicator of the quality of the generated images. 
We observed that DP2, CIAGAN and Fawkes-High showed the lowest average bias, indicating that the images generated by these GAN-based methods include faces. 

Non-GAN-based methods generally exhibited more bias, as they tend to modify faces more drastically. An exception was \emph{Fawkes}, which relies on imperceptible changes to the face and therefore has minimal impact on image quality. Bias was observed in \emph{Pixelation}, particularly against most demographics, including Black males and Asian females, with White males also affected in both BFW and DemogPairs. For \emph{DP-Snow} faces, \emph{MTCNN} showed severe bias against Black males in both BFW and DemogPairs, and against Black individuals in RFW. Finally, \emph{K-Same-Pixel-ArcFace} showed bias against Black males, Asian males, and females due to lower-quality faces.

\textbf{Summary:} MTCNN was the best detector with the least bias followed by SSD and RetinaFace. 
BFW was the worst performing dataset with the highest bias while CelebA was the best performing with the least bias. DP2 and CIAGAN showed high performance and lower bias than the original datasets on average. 
Bias was more pronounced among paired demographics, particularly against Black males, Black individuals, and males in general.

\subsection{Face Recognition Evaluation}
In our threat model, we assume the attacker can examine different FR models to choose the best for the re-identification attacks. Accordingly, we examined several FR models in DeepFace, including VGG-Face, Facenet, OpenFace, DeepID, Dlib, ArcFace, and DeepFace, on all the original datasets. Note that with \sys{}, any other face recognition can be implemented and tested. 
To measure the performance, we used the F1-score, AUC, and TPR at FPR=10\%. The choice of a relatively high FPR was due to the fact that at lower FPRs, such as 0.01\%, the TPR was too low. This will be further explained in the following sections. 
To calculate FPRs, faces were selected from within the same demographic groups. 

\begin{table}[t]
\centering
\caption{Evaluation of FR systems available on DeepFace.} 
\resizebox{0.75\columnwidth}{!}{%
 \begin{tabular}{c|c c c } 
  \hline
 \textbf{FR}& \textbf{F1-Score}&\textbf{AUC}&\textbf{TPR at FPR=0.1}\\
     \hline
   ArcFace&.87&.91&.84 \\
   FaceNet&.83&.9&.78 \\
   Dlib&.81&.89&.77 \\
   VGG-Face&.80&.87&.68 \\
   OpenFace&.66&.67&.29 \\
   DeepFace&.66&.63&.26 \\
    \hline
\end{tabular}
\label{table:FRs} 
}
\end{table} 

Table~\ref{table:FRs} presents the results, with ArcFace achieving the highest scores across all three metrics. However, the overall findings were unexpected, as they did not align with the performance reported in previous studies~\cite{baltruvsaitis2016openface, taigman2014deepface, king2009dlib, deng2019arcface}. 
To investigate the potential impact of testing datasets on the results, we repeated the experiments on the LFW dataset, which is the most commonly used dataset for evaluating these methods. 
Table~\ref{table:lfw} shows the results, using Cosine Similarity as the distance metric to compare face representations. In addition to reporting TPRs at FPR = 0.1, we also included TPRs at FPR = 0.05, as the thresholds for achieving the best F1-scores for the top models were close to this value. 
Interestingly, while the performances on the LFW dataset were significantly higher, supporting our hypothesis that the choice of the testing dataset is a contributing factor, the TPRs remained substantially lower than those reported in the original papers, $FPR=1e-4$. 
The discrepancies in the results may also stem from the implementation provided by DeepFace. As noted on their website~\cite{serengil2020lightface}, ``the numbers obtained by the experiments using DeepFace might differ due to variations in normalization and detection methods, as the actual pre-trained weights for some methods were not provided.''

\subsection{Verification Attacks with DeepFace}

Face verification was performed on all original and obfuscated datasets. Additionally, results for demographic pairs in BFW and DemogPairs were merged to analyze performance across different genders and races separately. In particular, for each method in the verification attacks, we present a summary of the results across three categories: 
\emph{Demographic Pairs}: This includes datasets with demographic pairs, specifically BFW and DemogPairs.
\emph{Race}: This includes RFW combined with a merged version of BFW and DemogPairs, grouped by race.
\emph{Gender}: This includes CelebA along with a merged version of BFW and DemogPairs, grouped by gender.
For some methods, including the original and GAN-based approaches, results are provided for all face recognition (FR) models across the four datasets. However, for brevity, results for the remaining methods are restricted to the ArcFace model on the BFW and DemogPairs datasets. 

\subsubsection{Average Results}

\textbf{Methods:} Based on Table~\ref{table:overall}, the average verification rate for the original datasets is 83\%. 
The highest overall success rate (OSR) was observed with Pixelation (93\%), while Fawkes had the lowest, at just 2\%. The high OSR for Pixelation can be attributed to the sufficiently large pixelation window, which effectively obscures the identity. 
Fawkes' poor obfuscation is explained in detail later. 
DP2 shows a high OSR, indicating it is an effective GAN-based obfuscation method with high-quality faces. The verification attack used here is a one-to-one comparison, with a chance accuracy of 50\%, and the attacker has access to the original photo, increasing the likelihood of recognition. DP1 has an average OSR of 69.5\%, higher than chance accuracy. Surprisingly, DP-Snow achieves an OSR of 58\%, higher than K-Same-Pixel-ArcFace (42\%), despite the latter heavily distorting the face, while DP-Snow only adds snow pixels that remain recognizable. 

\begin{table}
\centering
\caption{Verification rates and OSRs using ArcFace}
\label{table:verification_total_dataset_mean}
\begin{tabular}{lllll}
\hline
\multirow{2}{*}{\textbf{Method}}& \multicolumn{4}{c}{\textbf{Dataset}}\\ 

\cline{2-5}
& BFW & DemogPairs& RFW&CelebA\\ \hline
No-obfuscation&85.8&80.5&70.3&89.9\\
CIAGAN &46.0&48.9&48.6&31.3\\
DP2 &83.0&71.1&72.5&75.8\\
Fawkes-High&1.1&3.9&&\\
DP-Snow&64.8&48.2&&\\
Pixelation&93.7&91.6&&\\
K-Same-Pixel-ArcFace&27.2&78.6&&\\
\hline 
\end{tabular}
\end{table}

\textbf{Datasets:} 
Table~\ref{table:verification_total_dataset_mean} shows that CelebA has the highest average verification rate (55.3\%), compared to the other datasets, which range from 39\% to 48.6\%.
As expected, this dataset had the lowest OSRs for the three GAN-based methods, as being easier to recognize makes it harder to obfuscate. 
DP1 and DP2 had significantly higher OSRs for BFW compared to CIAGAN.

\subsubsection{Average Bias}

\textbf{Datasets.}
Table~\ref{table:verification_datasets-ratio} in Appendix shows the average bias for each method and dataset for the 5 values of $\epsilon$. Here, we summarize key findings. 

\textbf{Methods:}
Pixelation exhibited the lowest occurrence of bias, likely due to its high level of obfuscation from a large pixelating window. Interestingly, DP1 and DP2 also demonstrated lower bias, with zero bias towards any race in BFW across all $\epsilon$ values. In contrast, CIAGAN, along with the other methods, performed the worst in terms of bias across all datasets.

\subsubsection{Demographic Bias}

\textbf{Pairs:}
For the original datasets, ArcFace was biased against White Females and Asian Males in both BFW and DemogPairs, and against Asian Females in DemogPairs for $\epsilon \geq 0.1$. Fawkes, K-Same, Pixelation, and DP1 showed bias against Black Males in both datasets.
All methods, except CIAGAN, exhibited bias against Asian Males and Females in DemogPairs, indicating a broader trend of bias against Asians in both face recognition and obfuscation. These biases were observed for $\epsilon \geq 0.1$ in DP1, DP2, and Pixelation, and across all $\epsilon$ values for the other methods.

\textbf{Races:}

\emph{CIAGAN} was biased against Whites and Indians, \emph{DP1} was mostly fair but biased against Asians in RFW, and \emph{DP2} was most biased against Asians. Other methods showed bias against Blacks in BFW and DemogPairs, although Blacks had the highest TPRs in DemogPairs and RFW.
 
\textbf{Gender:}
ArcFace showed bias against females in DemogPairs and CelebA, while all obfuscation methods, except K-Same, were biased against males.

\textbf{Summary}
Black Males (and Blacks in general) had high verification rates but the lowest OSRs in most methods. Asians, especially Asian Males, faced bias in both recognition and obfuscation. Similarly, Males experienced a situation similar to Blacks regarding gender bias.

\textbf{Discussing Fawkes Results}

Fawkes introduces imperceptible changes to the face to alter its identity for machine recognition, while remaining recognizable to humans. 
The results reported by Szegedy et al.~\cite{shan2020fawkes} suggested that using a different FR model for attacks would lead to a slight decrease in obfuscation rates by a few percent. However, our findings show a much larger degradation, with obfuscation rates dropping by several tens of percent.
It is also suggested that future FR models could resist Fawkes if trained with cloaked faces generated by Fawkes. However, Radiya-Dixit et al.~\cite{radiya2021data} showed that adversarial examples are generally ineffective against future FR models, even without specific robustness training. While ArcFace was developed before Fawkes, some versions of ArcFace were released after Fawkes v1.0, which may explain why Fawkes is less effective against ArcFace.

\subsection{Verification Attacks by Face++}
We also used Face++ for verification attacks of the 3 GAN-based obfuscation methods. 
In Face++, confidence scores rather than the distance scores used in DeepFace, show how confident the system is about the faces being the same. 

We chose 70 as the optimal threshold~\cite{khorzooghi2022stylegan} to apply to the confidence scores to calculate VRs and OSRs.

\textbf{Average Results:}
Table~\ref{table:Verification_FPP_total_dataset_mean} shows that the verification rates for BFW and DemogPairs were nearly identical (89.5\% vs. 90.5\%), both higher than the average verification rate of 84\% for ArcFace. 
DP2 had the highest average OSRs compared with DP1 and CIAGAN with OSRs 97.2\% \& 91.3\% for BFW and DemogPairs compared with 97.2\% \& 88.4\% for DP1 and 82.7\% and 85.1\% for CIAGAN, indicating high performance of these methods in obfuscation and excellence of DeepPrivacy methods compared to CIAGAN.  
Overall, Face++ is a more effective face verification system since it gives both higher VRs and OSRs compared to Face++.

\textbf{Average Bias:}
 
\textbf{Datasets:} Bias occurred more frequently in DemogPairs across all demographics and methods compared to BFW, consistent with the results from DeepFace. 
\textbf{Methods:}
Similar to DeepFace results, CIAGAN exhibited the highest bias across all demographics and datasets, with 100\% bias for all $\epsilon$ values in gender, while other methods showed zero bias for most $\epsilon$ values. 
For DP1 and DP2, bias was generally similar to that in the original datasets. 

\textbf{Demographic Bias:}

\textbf{Pairs} 

DP1 and DP2 are unfair to White Females and Males for small $\epsilon$s, and CIAGAN was also against White Females the most for all $\epsilon$s.
\textbf{Race} 
Face++ is against Asians the most for verification. When applying on obfuscated faces, Whites will have the least unfair OSRs. All methods favor Asians the most.
\textbf{Gender}
CIAGAN is unfair to Females for all $\epsilon$s. DP1 and DP2 are also biased against Females in only DemogPairs for $\epsilon=0.02$

\textbf{Summary}
Face++ showed much higher rates in both original and obfuscated datasets, highlighting its superiority. Demographics with high verification rates had lower obfuscation success, with Asians/Whites experiencing the most in verification/obfuscation and vice versa. Gender exhibited minimal bias, with almost zero bias for most $\epsilon$ values in DP1 and DP2, while race and demographic pairs had higher biases, indicating that bias in these obfuscation methods is more influenced by race than gender.

\subsection{Identification} 

For identification, \sys{} obtains face descriptors for all original and obfuscated faces in BFW and DemogPairs, using ArcFace to generate the representations. In our experiments, 80\% of the face descriptors for each identity were used to train a Multi-SVM classifier, which employs multiple two-class SVMs. The assigned identity is determined by majority voting, with the identity having the highest number of wins being selected.

\textbf{Average Results:}
The average Identification Rate (shown in Table~\ref{table:overall}) is slightly higher while the average OSRs are slightly lower for S2, which is expected since training the models on a specific demographic allows for easier face distinction within that demographic. This results in a higher identification rate and lower OSRs in $S_2$. However, this trend does not hold for CIAGAN, as the demographics of the obfuscated faces often differ from those of the original faces, which could be more pronounced in the case of CIAGAN.

OSRs for identification attacks are higher than for verification attacks for two reasons: 1) the original photo is not used in identification training, leading to less similarity, and 2) the larger number of classes in identification increases the chance of misprediction. For the original datasets, identification and verification rates are similar, showing ArcFace's generalization capability despite high FPRs. Regarding obfuscation methods, all but DP-Snow (79-73\%) and Fawkes (31-37\%) have OSRs above 90\%, which is notable.

\textbf{Datasets:} Table~\ref{Table:Identification_datasets-ratio} shows the results.
On average, BFW had less bias for both scenarios. 

\textbf{Scenarios:} $S_2$ generally exhibited more bias than $S_1$ due to lowers OSRs. Accordingly, the original datasets had less bias in $S_2$ due to higher IRs. 
\textbf{Methods:}
All methods except for Fawkes and DP-Snow had less bias than the original datasets due to higher OSRs than IRs. These methods had zero bias for almost all $\epsilon \geq 0.1$.

\textbf{Demographic Bias}
We only discuss $S_2$ due to being a more powerful attack and having more bias. We observed bias everywhere. 

\textit{CIAGAN} was biased against Asian males and Whites in DemogPairs, and Whites and Indian males in BFW. \textit{DP1} and \textit{DP2} were biased against White males and other males in both datasets, with \textit{DP1} also affecting Black Females.
\textit{Fawkes} showed high, inconsistent bias, particularly against White and Indian males in BFW and Asians in DemogPairs. \textit{DP-Snow} was biased against Whites in DemogPairs and White/Indian Males in BFW. \textit{Pixelation} had lower bias against Whites in BFW and Asian Males/Blacks in DemogPairs. \textit{K-Same} was biased only against White Females in DemogPairs and all races except Asians in BFW.

\subsection{Utility Analysis} 
We used \sys{} to test whether GAN-based methods preserve different attributes. The DeepFace library measures four attributes: age, gender, race, and emotion. We compared these attributes between obfuscated faces and their original versions. Age is given as a number, gender as Woman or Man, emotion as angry, disgust, fear, happy, sad, surprise, or neutral, and race as Asian, Indian, Black, White, Middle Eastern, or Latino Hispanic.

\textbf{Gender Analysis: }
Table~\ref{table:attribute_datasets-ratio} show Average Bias.
DP2 had the higher average Gender Preserving Rate (84\%) and CIAGAN had the lowest (77\%). Regarding datasets, CelebA had the highest average rate of 81\% while BFW had the lowest(72\%)
\textbf{Average Bias:} 
As expected, CelebA showed the lowest bias, with zero bias for larger $\epsilon$, while BFW showed the highest bias across all $\epsilon$. All three methods exhibited bias, with DP1 and DP2 showing higher bias than CIAGAN, which had zero bias for CelebA. 
\textbf{Demographic Bias}
Bias existed for almost all $\epsilon$. 
For all the methods, Females, Asians, and mostly Asian Females suffered the most from lower Gender PRs in the related datasets. Black Males and Blacks on the other hand had higher Gender PRs.

\textbf{Emotion Analysis:} 
\emph{Average Results: } Emotion Preserving rates were not as high as Gender, with 45\%, 42\%, and 39\% for CIAGAN, DP1, and DP2. Again CelebA had the highest average rate while BFW had the lowest.
\textbf{Average Bias:}
The methods had a similar bias as for Gender, with this difference that CelebA and CIAGAN also showed a high bias for most $\epsilon$s. 
\textbf{Demographic Bias:}
Despite Gender, Males suffered the most, especially in DP1 and DP2. Regarding Race, mostly Asians followed by Blacks unfairly had lower Emotion Preserving Rates.
\textbf{Race Analysis:}
\emph{Average Results:}
Race PRs were slightly higher the Emotion Rates for the three methods, being around 50\%.
Again, CelebA had the highest average rate, 62\%, vs. BFW having around 30\%.
\textbf{Average Bias:}
We observed bias in each method and dataset for almost all $\epsilon$s except for larger $\epsilon$s for CelebA in DP1 and DP2.
\textbf{Demographic Bias:}
Race PRs were highly biased against Black Males and Females in both BFW and DemogPairs and highly favored White Females. In RFW, Whites also had unfairly higher rates than the others. In CelebA, the rates were high against Males in all three methods.

\textbf{Age Analysis:}
\emph{Average Results:}
Age PRs were similar for all the methods and datasets, with a minimum of 81\% for DP2-BFW and a maximum of 86\% for CelebA for all the methods.
\textbf{Average Bias:}
Due to similar average rates, less bias was observed in Age Preserving Rates, with being zero for all $\epsilon \geq 0.15$ for all cases.
\textbf{Demographic Bias:}
Although less bias is observed regarding age, the bias was observed mostly against Males, especially White Males for $\epsilon \leq 0.05$

\textbf{Summary}
Regarding attribute preserving rates, Race had the highest bias while age showed the lowest. Regarding datasets, CelebA had the lowest bias. 
Gender and Emotion Preserving Rates were most against Asians while Race Preserving Rates were against Blacks. Excluding Gender, the other preserving rates were mostly against Males.

\subsection{Face++ Attribute Analysis}
We consider emotion, age, gender, and headpose. Face++ does not provide race estimation. The preserving rates were calculated as in DeepFace.
For the pose, three angles: yaw, pitch, and roll angles are estimated, which are modeled as a three-dimensional vector so that cosine similarity can be used to measure pose vector similarities. Tables~\ref{table:Attribute_FPP_total_dataset_means} and \ref{table:Attribute_FPP_datasets-ratio}
show dataset means and the average bias for datasets, respectively.

\textbf{Gender:}
DP1 and DP2 had the same average gender preserving rates which were higher than CIAGAN (82\% vs 66\%).
Rates regarding DemogPairs were also higher than BFW, especially for DP1 and DP2 (85\% vs 74\%). 
\textbf{Average Bias}
Bias existed for all $\epsilon$ for all methods and datasets. CIAGAN had a higher bias, especially for higher $\epsilon$s.
\textbf{Demographic Bias}
Asian Males suffered the most in all cases with a bias of 100\% for most $\epsilon$. 
All Males in CIAGAN also had unfairly lower rates.

\textbf{Emotion:}
Emotion-preserving rates were lower with an average of around 45\%. CIAGAN had rates across datasets while DP1 and DP2 performed better for DemogPairs.
\textbf{Average Bias:}
The bias was similar to that of gender.
\textbf{Demographic Bias:}
Males usually suffered the most, along with Asian Females in CIAGAN-DemogPairs. 

\textbf{Age:}
Average Age Preserving Rates were quite high, around 74\% for BFW and 77\% for DemogPairs. They were slightly higher for DP1 and DP2.
\textbf{Average Bias:}
Due to the rates being similar, not much bias was observed. DP1 and DP2 showed bias only for the smallest $\epsilon$, while CIAGAN showed more bias, especially for BFW in all $\epsilon$. 
\textbf{Demographic Bias:}
CIAGAN was biased against Males, especially White Males. DP1 and DP2 were mostly against Asian Males for $\epsilon=.02$

\textbf{Head Pose:}
The average rates were more consistent for CIAGAN (79.5 and 77.4\%) across the datasets while they were much lower for BFW for DP1 and DP2. 63.9\% and 49.2\% vs 87.4\% and 82.6\%.
\textbf{Average Bias}
The bias was similar for the methods, mostly for smaller $\epsilon$s.
\textbf{Demographic Bias}
Head Pose Preserving Rates were mostly against Black Males while they favored Asian Females the most.

\textbf{Summary:}
Generally, CIAGAN showed smaller rates and higher bias. 
Age and Head pose showed lower bias than Gender and Emotion, which also showed the lowest rates. The rates were mostly against Males.

\subsection{Visualization and Source of Bias}
Among various face obfuscation methods, we focus on visualizing the Fawkes and CIAGAN algorithms as they have lower obfuscation success rates when compared to other algorithms. Additionally, they often showed more bias towards certain demographics compared to other methods.

To understand how these modifications vary across different demographics, we created two types of visualizations for each algorithm: one analyzing the distribution across races by combining data from both genders for each racial group, and another examining gender-based patterns by combining data from all races for each gender.

\textbf{Fawkes Feature Distribution Analysis}
Figures~\ref{fig:viz_fawkes_races} and ~\ref{fig:viz_fawkes_genders} the number of images for different focus features for race and gender respectively. 

Across different racial groups, we observed \emph{Indian} and \emph{Black} individuals consistently showed the highest intensities, in the nose bridge 
and forehead. 
Asian individuals show variable patterns, with high focus in nose regions 
but lower intensities in other areas. White individuals exhibit low intensities across most features, and the focus is mainly on central facial regions like nose bridge and forehead.
Our gender-based analysis of Fawkes revealed striking disparities in how the algorithm processes male and female subjects. Female subjects show extremely high focus on the nose bridge
and nose tip, with significantly higher values compared to males in these regions. Male subjects in contrast show high focus on forehead 
and nose
along with right cheek. 

\textbf{CIAGAN Feature Distribution Analysis}
We show the feature distribution between races using CIAGAN in Figure~\ref{fig:viz_ciagan_races}. 

CIAGAN demonstrated a different approach to feature focus compared to Fawkes. Although still keeping a focus on central regions, it also showed increased attention to eye regions and peripheral features. \emph{Black} subjects show slightly higher values in the nose tip 
, while Indian subjects demonstrate higher modifications in the forehead region 
. Asian and White subjects maintain more moderate intensities, typically ranging from 800-1000 images for central features. 
The gender-based analysis of CIAGAN revealed smaller disparities between male and female subjects compared to Fawkes. Female subjects show the highest attention to nose tip 
and nose bridge.
Male subjects demonstrate a strong focus on the forehead 
and nose.
Unlike Fawkes, CIAGAN maintains more consistent attention to secondary features, with eye regions, cheeks, and jaw areas receiving moderate modifications
across both genders. The distribution pattern shows a more balanced approach between genders, with smaller gaps between male and female modification intensities.

\textbf{Summary:} Both Fawkes and CIAGAN demonstrate distinct patterns in their treatment of facial features across demographics. When races are analyzed, Indian and Black subjects received highest intensities on central facial features like nose bridge and forehead. Across different genders, both systems focus more on the nose bridge and nose tip for female subjects, whereas the forehead area is for male subjects. The systems consistently prioritize central facial features (nose bridge, nose tip, forehead) across all demographic groups, with peripheral features (eyes, cheeks, chin) receiving moderate to lower intensities.

\subsection{Correlation Analysis}

Our analysis revealed notably low correlation coefficients across all demographic groups, as shown in Table~\ref{tab:correlations}. For Fawkes, correlations ranged from 0.08 to 0.13, with Asian males showing the highest correlation (0.13) and Black females showing the lowest (0.08). CIAGAN exhibited even lower correlations, ranging from 0.06 to 0.10, following a similar demographic pattern.

\begin{table}[t]
\centering
\caption{Correlation Coefficients Between Obfuscation and Recognition Model Focus Patterns Using BFW Dataset}
\label{tab:correlations}
\begin{tabular}{lcc}
\hline
Demographic Group & Fawkes & CIAGAN \\
\hline
White Female & 0.09 & 0.07 \\
White Male   & 0.11 & 0.08 \\
Black Female & 0.08 & 0.06 \\
Black Male   & 0.10 & 0.07 \\
Asian Female & 0.12 & 0.09 \\
Asian Male   & 0.13 & 0.10 \\
Indian Female & 0.11 & 0.08 \\
Indian Male  & 0.12 & 0.09 \\
\hline
\end{tabular}
\end{table}

These low correlations reveal a fundamental mismatch between obfuscation methods and recognition systems. While obfuscation targets specific features (primarily the nose area), recognition systems analyze faces holistically, explaining their resilience to targeted obfuscation. Correlation variations across demographics (higher for Asians, lower for Black and White subjects) align with the OSR patterns observed in verification results. This suggests two critical improvements needed for face obfuscation methods: (1) implementing more holistic changes to effectively counter recognition systems, and (2) ensuring uniform changes across all demographics to eliminate bias. Particularly, adversarial methods like Fawkes would benefit from supervised approaches that address both these aspects.

\section{Discussion, Limitations, and Future Work}
One finding from the results is that generally, higher rates are associated with lower bias. To reduce bias, the rates should be increased. However, performance differences become more apparent at higher rates, while fairness formulas tend to show more bias in lower-performance systems with similar rate differences. 
Thus, bias estimation should align with performance rates, with experts selecting lower $\epsilon$s for systems with higher rates. 
MTCNN proved to be the optimal detection method, showing best performance and lowest bias, making it recommended for face recognition and obfuscation methods relying on face detection.
In privacy preservation methods, Pixelation performed best despite quality sacrifices due to large pixelation windows. GAN-based methods also showed strong results – DP2 achieved high privacy and detection rates with low bias, while CIAGAN showed higher bias. Bias patterns remained consistent across datasets, except for BFW's occasional variations due to quality issues. Notably, methods consistently showed bias against Males, particularly Black Males.

Feature focus analysis revealed that Fawkes and CIAGAN prioritize changing different face regions for different demographics. This suggests that obfuscation methods should aim for similar changes across demographics, potentially by adding an average heatmap mask constraint to their cost function. This approach, similar to face recognition attacks, would help target key feature areas more effectively, improving their ability to evade verification and identification

Fawkes failed in face obfuscation because it relied on exploiting specific weaknesses through imperceptible changes. However, robust face recognition tools like ArcFace, which uses the entire face for identification, proved resistant to these subtle changes, as indicated by the results from the correlation analysis section. As a limitation, we only performed utility-preserving analyses rather than the whole utility concept and its trade-off with privacy. The reason was that we focused on specific attributes that could be important for privacy-preserving applications.

\section{Conclusion}
In this work, we presented a comprehensive framework \sys{} to assess the adversarial robustness and fairness of various face obfuscation methods. The framework introduced encompasses various set of modules, that includes 4 benchmark dataset, various face detection and recognition algorithms, adversarial models, utility detection models, and fairness metrics. 
We also performed visualization and focus feature comparisons to find out that face obfuscation methods should focus on those feature regions that are important for robust face recognition systems. 
We also presented using different thresholds especially smaller ones for high-performance systems to reach a better perspective toward actual bias.

\bibliographystyle{IEEEtran}
\bibliography{refs}

\appendix
\begin{table}[t]
\centering
\caption{Evaluation of DeepFace's FR systems on LFW } 
\label{table:lfw}
\resizebox{0.9\columnwidth}{!}{%
 \begin{tabular}{c|c c c c } 
  \hline
 \textbf{FR}& \textbf{F1-Score}&\textbf{AUC}&\textbf{TPR|FPR=0.1}&\textbf{TPR|FPR=0.05}\\
     \hline
   FaceNet&.95&.98&.98&.96\\
   ArcFace&.95&.98&.98&.96 \\
   VGG-Face&.89&.96&.89&.84 \\
   Dlib&.85&.95&.83&.72\\
   OpenFace&.66&.67&.31&.29\\
   DeepFace&.66&.63&.33&.26 \\
   DeepID&.60&.66&.24&.16\\
    \hline
\end{tabular}
}
\end{table}
\begin{table}[t]
\centering
\caption{Average Attribute Preserving rates of DeepFace} 
\resizebox{.70\linewidth}{!}{%
 \begin{tabular}{c|cccc} 
  \hline
 \textbf{Algorithms}& \textbf{Gender}&\textbf{Race}&\textbf{Emotion}&\textbf{Age}\\
     \hline
   CIAGAN&77&50&45&85 \\
   DP1&83&52&42&85\\
   DP2&84&52&39&84\\
    \hline
\end{tabular}
\label{table:overall-utility} 
}
\end{table} 
\begin{figure}[h!]
    \centering
    \includegraphics[width=0.9\linewidth]{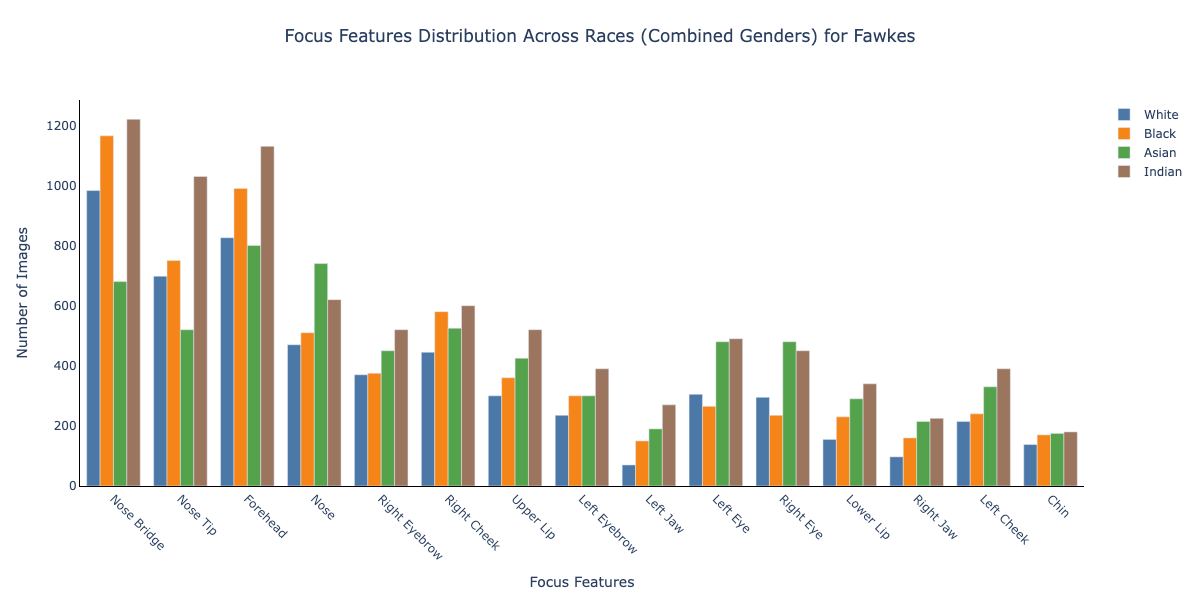}
    \caption{Focus Features Distribution Across Races (Combined Genders) for Fawkes}
    \label{fig:viz_fawkes_races}
\end{figure}

\begin{figure}[h!]
    \centering
    \includegraphics[width=0.9\linewidth]{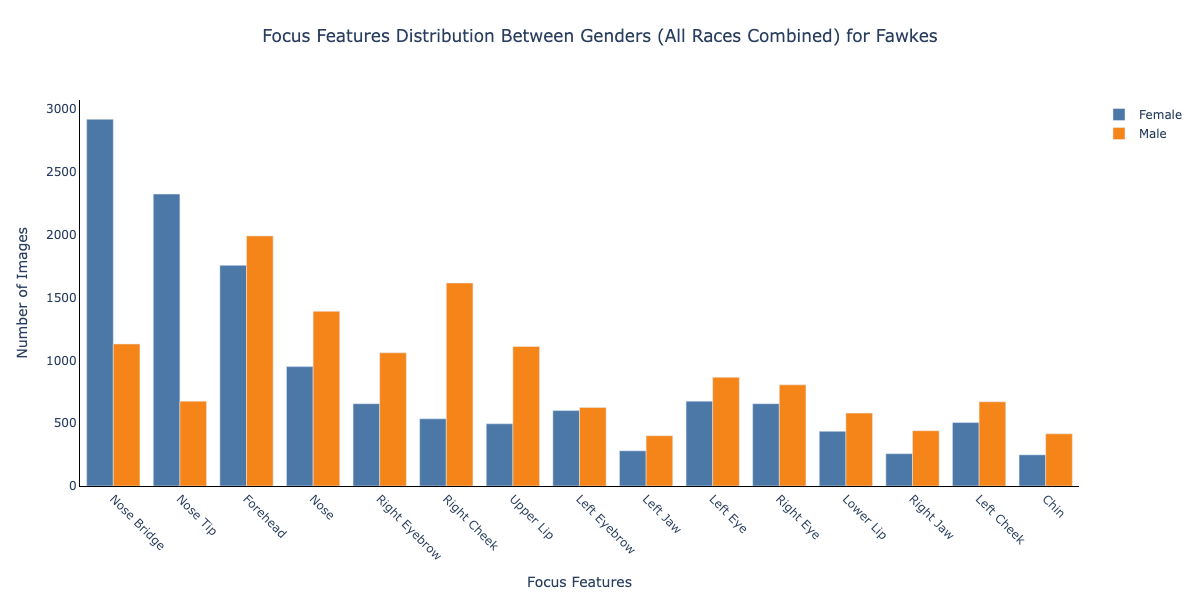}
    \caption{Focus Features Distribution Between Genders (All Races Combined) for Fawkes}
    \label{fig:viz_fawkes_genders}
\end{figure}

\begin{figure}[h!]
    \centering
    \includegraphics[width=0.9\linewidth]{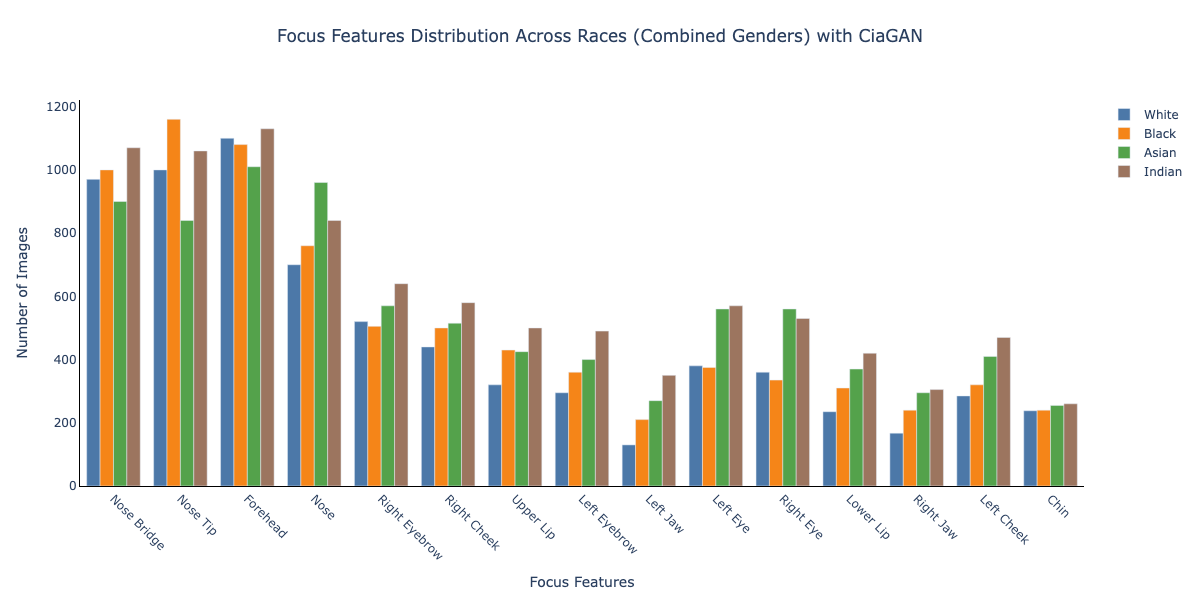}
    \caption{Focus Features Distribution Across Races (Combined Genders) for CiaGAN}
    \label{fig:viz_ciagan_races}
\end{figure}


\begin{table}[t]
\centering
\caption{Verification rates and OSRs for Face++} 
\resizebox{0.5\columnwidth}{!}{%
 \begin{tabular}{c|c c } 
  \hline
  \textbf{Method}& \textbf{Datasets}&\\
  \hline
 \textbf{Original}& \textbf{BFW}&\textbf{DemogPairs}\\
     \hline
   pairs&89.9&90.5 \\
   race&89.9&90.5 \\
  Gender&89.9&90.5 \\
  \hline
  \textbf{DP1}& \textbf{BFW}&\textbf{DemogPairs}\\
  \hline
  pairs&97.2&88.4 \\
   race&97.2&88.4 \\
  Gender&97.2&88.4 \\
    \hline
    \textbf{DP2}& \textbf{BFW}&\textbf{DemogPairs}\\
  \hline
  pairs&97.2&91.3 \\
   race&97.2&91.3 \\
  Gender&97.2&91.3 \\
  \hline
  \textbf{CIAGAN}& \textbf{BFW}&\textbf{DemogPairs}\\
\hline
pairs&82.7&85.1 \\
   race&82.7&85.1 \\
  Gender&82.7&85.1 \\
\hline
\end{tabular}
\label{table:Verification_FPP_total_dataset_mean} 
}
\end{table} 

\begin{table}[t]
\centering
\caption{Average Bias for Identification in DeepFace for $\epsilon=[0.02, 0.05, .1, .15, .2]$ } 
\resizebox{\columnwidth}{!}{%
 \begin{tabular}{c|c c } 
  \hline
  \textbf{$S_1$}& \textbf{Datasets}&\\
  \hline
 & \textbf{BFW}&\textbf{DemogPairs}\\
     \hline
Original&[82, 64, 32, 7, 0]&[93, 67, 40, 13, 0]\\
CIAGAN&[50, 0, 0, 0, 0]&[60, 27, 0, 0, 0]\\
DP1&[4, 0, 0, 0, 0]&[67, 20, 0, 0, 0]\\
DP2&[0, 0, 0, 0, 0]&[60, 27, 0, 0, 0]\\
Fawkes-High&[96, 96, 86, 79, 71]&[100, 93, 93, 93, 73]\\
DP-Snow&[89, 57, 29, 7, 0]&[87, 67, 20, 13, 0]\\
Pixelation&[21, 0, 0, 0, 0]&[0, 0, 0, 0, 0]\\
K-Same-Pixel-ArcFace&[46, 4, 0, 0, 0]&[13, 0, 0, 0, 0]\\
means&[48 28 18 12  9]&[60 38 19 15  9]\\
  \hline
  \textbf{$S_2$}& \\
  \hline
Original&[82, 64, 29, 7, 0]&[87, 60, 33, 33, 13]\\
CIAGAN&[79, 43, 14, 0, 0]&[73, 33, 0, 0, 0]\\
DP1&[29, 0, 0, 0, 0]&[73, 47, 13, 0, 0]\\
DP2&[46, 0, 0, 0, 0]&[73, 33, 0, 0, 0]\\
Fawkes-High&[96, 96, 89, 86, 75]&[93, 93, 93, 93, 80]\\
DP-Snow&[93, 61, 32, 11, 0]&[73, 40, 33, 27, 0]\\
Pixelation&[29, 0, 0, 0, 0]&[60, 20, 0, 0, 0]\\
K-Same-Pixel-ArcFace&[54, 4, 0, 0, 0]&[13, 0, 0, 0, 0]\\
means&[64 34 20 13  9]&[68 41 22 19 12]\\
\hline
\end{tabular}
\label{Table:Identification_datasets-ratio} 
}
\end{table} 

\begin{table}[t]
\centering
\caption{Attribute Preserving Rates for Face++} 
\resizebox{0.5\columnwidth}{!}{%
 \begin{tabular}{c|c c } 
  \hline
  \textbf{}& \textbf{Datasets}&\\
  \hline
 \textbf{gender}& \textbf{BFW}&\textbf{DemogPairs}\\
     \hline
   DP1&71.4&85.5 \\
   DP2&76.5&84.4\\
  CIAGAN&67.1&65.0\\
  \hline
  \textbf{emotion}& &\\
  \hline
  DP1&34.2&48.9 \\
   DP2&36.7&48.6\\
  CIAGAN&47.3&45.9\\
    \hline
\textbf{age}& &\\
\hline
DP1&74.2&78.2\\
   DP2&74.8&78.3\\
  CIAGAN&73.6&75.5\\
\hline
\textbf{headpose}& &\\
\hline
DP1&63.9&87.4\\
   DP2&49.2&82.6\\
  CIAGAN&79.5&77.4\\
\hline
\end{tabular}
\label{table:Attribute_FPP_total_dataset_means} 
}
\end{table} 

\begin{table}
\centering
\caption{Average Bias of Attribute Preserving Rates for $\epsilon=[.02, .05, .1, .15, .2]$ in Face++ }
\resizebox{\columnwidth}{!}{%
\label{table:Attribute_FPP_datasets-ratio}
\begin{tabular}{p{0.08\linewidth}|lll}
\hline
\multirow{1}{*}{\textbf{}}& \multicolumn{3}{c}{\textbf{Dataset}}\\ 
\cline{1-4}
\textbf{gender}& BFW & DemogPairs&mean \\ \hline
DP1&[75, 61, 46, 18, 18]&[93, 67, 40, 27, 13]&[84, 64, 43, 22, 16]\\
DP2&[82, 71, 46, 39, 25]&[73, 53, 53, 33, 0]&[78, 62, 50, 36, 12]\\
CIAGAN&[93, 86, 75, 64, 64]&[100, 87, 73, 60, 60]&[96, 86, 74, 62, 62]\\
mean&[83, 73, 56, 40, 36]&[89, 69, 55, 40, 24]&\\
\hline
\textbf{emotion}& BFW & DemogPairs&mean \\ 
\hline
DP1&[93, 79, 64, 57, 39]&[87, 87, 80, 53, 53]&[90, 83, 72, 55, 46]\\
DP2&[82, 79, 43, 21, 0]&[100, 93, 67, 53, 27]&[91, 86, 55, 37, 14]\\
CIAGAN&[86, 64, 36, 7, 4]&[87, 60, 47, 7, 7]&[86, 62, 42, 7, 6]\\
mean&[87, 74, 48, 28, 14]&[91, 80, 65, 38, 29]&\\
\hline
\textbf{age}& BFW & DemogPairs&mean \\
\hline
DP1&[43, 0, 0, 0, 0]&[47, 0, 0, 0, 0]&[45, 0, 0, 0, 0]\\
DP2&[39, 0, 0, 0, 0]&[40, 0, 0, 0, 0]&[40, 0, 0, 0, 0]\\
CIAGAN&[82, 68, 36, 11, 7]&[73, 47, 0, 0, 0]&[78, 58, 18, 6, 4]\\
mean&[55, 23, 12, 4, 2]&[53, 16, 0, 0, 0]&\\
\hline
\textbf{pose}& BFW & DemogPairs&mean \\
\hline
DP1&[89, 68, 50, 36, 25]&[73, 33, 7, 0, 0]&[81, 50, 28, 18, 12]\\
DP2&[93, 75, 50, 25, 11]&[60, 20, 0, 0, 0]&[76, 48, 25, 12, 6]\\
CIAGAN&[82, 46, 21, 0, 0]&[73, 33, 33, 27, 0]&[78, 40, 27, 14, 0]\\
mean&[88, 63, 40, 20, 12]&[69, 29, 13, 9, 0]&\\
\hline

\end{tabular}
}
\end{table}

\begin{table}[t]
\centering
\caption{Attribute Preserving Rates for DeepFace}
\resizebox{0.8\columnwidth}{!}{%
\label{table:Attribute_total_dataset_mean}
\begin{tabular}{p{0.15\linewidth}|lllll}
\hline
\multirow{1}{*}{\textbf{}}& \multicolumn{5}{c}{\textbf{Dataset}}\\ 
\cline{1-6}
\textbf{gender}& BFW &RFW& DemogPairs&CelebA&mean\\ \hline
CIAGAN&72&77&71&78&77\\
DP1&69&82&79&88&83\\
DP2&75&82&79&87&84\\
\hline
\textbf{emotion}& BFW &RFW& DemogPairs&CelebA&mean\\\hline
CIAGAN&42&39&38&49&45\\
DP1&28&35&42&49&42\\
DP2&26&34&39&45&39\\
\hline
\textbf{race}& BFW &RFW& DemogPairs&CelebA&mean\\\hline
CIAGAN&30&35&27&62&50\\
DP1&30&45&44&61&52\\
DP2&31&43&44&64&52\\
\hline
\textbf{age}&BFW &RFW& DemogPairs&CelebA&mean \\\hline

CIAGAN&86&84&86&86&85\\
DP1&84&84&86&86&85\\
DP2&81&84&85&86&84\\
\hline 
\end{tabular}
}
\end{table}

\begin{table*}
\centering
\caption{Average Bias for Attribute Preserving Rates for DeepFace for $\epsilon=[.02, .05, .1, .15, .2]$}
\resizebox{0.9\linewidth}{!}{%
\label{table:attribute_datasets-ratio}
\begin{tabular}{p{0.15\linewidth}|lllll}
\hline
\multirow{1}{*}{\textbf{}}& \multicolumn{5}{c}{\textbf{Dataset}}\\ 
\cline{1-6}
\textbf{gender}& BFW &RFW& DemogPairs&CelebA&mean\\ \hline
CIAGAN&[89, 82, 54, 32, 18]&[100, 83, 67, 50, 33]&[87, 87, 60, 47, 40]&[0, 0, 0, 0, 0]&[69, 63, 45, 32, 23]\\
DP1&[96, 89, 71, 68, 61]&[83, 83, 83, 17, 17]&	[93, 93, 87, 60, 47]&[100, 100, 100, 0, 0]	&[93, 91, 85, 36, 31]\\
DP2&[93, 79, 71, 68, 57]&[83, 83, 83, 17, 17]&	[100, 87, 80, 60, 53]&[100, 100, 100, 0, 0]	&[94, 87, 84, 36, 32]\\
mean&[93, 83, 65, 56, 45]&[89, 83, 78, 28, 22]&[93, 89, 76, 56, 47]&[67, 67, 67, 0, 0]&\\
\hline
\textbf{emotion}& BFW &RFW& DemogPairs&CelebA&mean\\\hline
CIAGAN&[89, 75, 50, 46, 39]&[100, 83, 33, 17, 0]&[87, 87, 33, 13, 7]&[100, 0, 0, 0, 0]&[94, 61, 29, 19, 12]\\
DP1&[89, 75, 57, 57, 57]&[100, 50, 17, 0, 0]&[93, 93, 67, 67, 47]&[100, 100, 100, 100, 0]&[96, 80, 60, 56, 26]\\
DP2&[96, 75, 57, 39, 14]&[67, 50, 50, 50, 0]&[100, 93, 67, 60, 53]&[100, 100, 100, 100, 100]&[91, 80, 68, 62, 42]\\
mean&[91, 75, 55, 47, 37]&[89, 61, 33, 22, 0]&[93, 91, 56, 47, 36]&[100, 67, 67, 67, 33]&\\
\hline
\textbf{race}& BFW &RFW& DemogPairs&CelebA&mean\\\hline
CIAGAN&[100, 96, 96, 89, 89]&[100, 100, 100, 100, 100]&[100, 100, 93, 93, 87]&[100, 100, 100, 100, 100]&[100, 99, 97, 96, 94]\\
DP1&[100, 96, 93, 86, 79]&[100, 100, 83, 83, 67]&[100, 100, 93, 93, 73]&[100, 100, 100, 0, 0]&[100, 99, 92, 66, 55]\\
DP2&[100, 100, 96, 89, 86]&[100, 100, 100, 100, 83]&[93, 93, 93, 87, 73]&[100, 100, 0, 0, 0]&[98, 98, 72, 69, 60]\\
mean&[100, 97, 95, 88, 85]&[100, 100, 94, 94, 83]&[98, 98, 93, 91, 78]&[100, 100, 67, 33, 33]&\\
\hline
\textbf{age}&BFW &RFW& DemogPairs&CelebA&mean \\\hline

CIAGAN&[86, 46, 11, 0, 0]&[83, 33, 0, 0, 0]&[67, 20, 0, 0, 0]&[100, 0, 0, 0, 0]&[84, 25, 3, 0, 0]\\
DP1&[68, 36, 0, 0, 0]&[0, 0, 0, 0, 0]&[60, 0, 0, 0, 0]&[100, 0, 0, 0, 0]&[57, 9, 0, 0, 0]\\
DP2&[64, 14, 0, 0, 0]&[17, 0, 0, 0, 0]&[60, 0, 0, 0, 0]&[100, 0, 0, 0, 0]&[60, 4, 0, 0, 0]\\
mean&[73, 32, 4, 0, 0]&[33, 11, 0, 0, 0]&[62, 7, 0, 0, 0]&[100, 0, 0, 0, 0]&\\
\hline 
\end{tabular}
}
\end{table*}

\begin{table*}
\centering
\caption{Detection Datasets Ratio results. Note that the results are shown for epsilon 0.2, 0.15, 0.1, 0.05, and 0.02 respectively from left to right. The higher the rate, the higher the bias occurrence between demographics.}
\resizebox{0.9\linewidth}{!}{%
\label{table:detection_datasets-ratio} 
\begin{tabular}{p{0.16\linewidth}lllll}
\hline
\multirow{1}{*}{\textbf{Method}}& \multicolumn{5}{c}{\textbf{Dataset}}\\ 
\cline{1-6}
\textbf{Original}& BFW & RFW& DemogPairs&CelebA& mean \\ \hline
opencv & [93, 68, 46, 32, 25] & [83, 50, 50, 50, 50]&[87, 80, 60, 40, 33]&[100, 0, 0, 0, 0]&[91.0, 50.0, 39.0, 30.0, 27.0]\\
ssd & [100, 96, 96, 93, 89]&[0, 0, 0, 0, 0]&[0, 0, 0, 0, 0]&[0, 0, 0, 0, 0]&[25.0, 24.0, 24.0, 23.0, 22.0]\\
dlib&[86, 54, 14, 4, 0]&[50, 50, 0, 0, 0]&[73, 40, 7, 0, 0]&[0, 0, 0, 0, 0]&[52.0, 36.0, 5.0, 1.0, 0.0]\\
retinaface & [96, 89, 71, 61, 32]&[0, 0, 0, 0, 0]&[0, 0, 0, 0, 0]&[0, 0, 0, 0, 0]&[24.0, 22.0, 18.0, 15.0, 8.0]\\
mtcnn&[0, 0, 0, 0, 0]&[0, 0, 0, 0, 0]&[0, 0, 0, 0, 0]&[0, 0, 0, 0, 0]&[0, 0, 0, 0, 0]\\
 \hline 
& \multicolumn{5}{c}{\textbf{mtcnn}}\\ 
  \cline{1-6}
\textbf{CIAGAN}&[0, 0, 0, 0, 0]&[0, 0, 0, 0, 0]&[0, 0, 0, 0, 0]&[0, 0, 0, 0, 0]&[0, 0, 0, 0, 0]\\
\textbf{DP1}&[86, 54, 25, 14, 4]&[0, 0, 0, 0, 0]&[7, 0, 0, 0, 0]&[0, 0, 0, 0, 0]&[23.0, 14.0, 6.0, 4.0, 1.0]\\
\textbf{DP2}&[0, 0, 0, 0, 0]&[0, 0, 0, 0, 0]&[0, 0, 0, 0, 0]&[0, 0, 0, 0, 0]&[0, 0, 0, 0, 0]\\
\textbf{Fawkes-High}&[0, 0, 0, 0, 0]&[0, 0, 0, 0, 0]&&&[0, 0, 0, 0, 0]\\
\textbf{DP-Snow}&[75, 46, 46, 25, 25]&[50, 33, 0, 0, 0]&[87, 47, 33, 27, 0]&[0, 0, 0, 0, 0]&[53.0, 32.0, 20.0, 13.0, 6.0]\\
\textbf{Pixelation}&[93, 86, 61, 36, 18]&[100, 100, 100, 87, 87]&&&[96.0, 93.0, 80.0, 62.0, 52.0]\\
\textbf{K-Same-Pixel-ArcFace}&[0, 0, 0, 0, 0]&[100, 93, 93, 87, 73]&&&[50.0, 46.0, 46.0, 44.0, 36.0]\\

\hline
\end{tabular}
}
\end{table*}

\begin{table*}
\centering
\caption{Verification datasets ratio results }
\resizebox{0.9\linewidth}{!}{%
\label{table:verification_datasets-ratio}
\begin{tabular}{p{0.15\linewidth}|llll}
\hline
\multirow{1}{*}{\textbf{}}& \multicolumn{4}{c}{\textbf{Dataset}}\\ 
\cline{1-5}
\textbf{pairs}& BFW & DemogPairs&mean& \\ \hline
Original&[50, 11, 0, 0, 0]&[87, 47, 13, 0, 0]&[68.0, 29.0, 6.0, 0.0, 0.0]&\\
CIAGAN&[100, 96, 93, 82, 71]&[93, 93, 80, 60, 53]&[96.0, 94.0, 86.0, 71.0, 62.0]&\\
DP1&[57, 11, 0, 0, 0]&[87, 87, 60, 40, 13]&[72.0, 49.0, 30.0, 20.0, 6.0]&\\
DP2&[71, 54, 11, 0, 0]&[93, 80, 53, 33, 13]&[82.0, 67.0, 32.0, 16.0, 6.0]&\\
Fawkes-High&[100, 100, 89, 89, 86]&[93, 93, 73, 73, 73]&[96.0, 96.0, 81.0, 81.0, 80.0]&\\
DP-Snow&[96, 75, 50, 25, 7]&[100, 93, 67, 33, 20]&[98.0, 84.0, 58.0, 29.0, 14.0]&\\
Pixelation&[61, 18, 0, 0, 0]&[67, 40, 13, 0, 0]&[64.0, 29.0, 6.0, 0.0, 0.0]&\\
K-Same-Pixel-ArcFace&[96, 96, 96, 86, 71]&[87, 80, 67, 53, 40]&[92.0, 88.0, 82.0, 70.0, 56.0]&\\
mean&[79.0, 58.0, 42.0, 35.0, 29.0]&[88.0, 77.0, 53.0, 36.0, 26.0]&&\\
\hline
\textbf{race}& BFW & DemogPairs&RFW&mean \\ 
\hline
Original&[50, 0, 0, 0, 0]&[100, 67, 0, 0, 0]&	[100, 67, 50, 50, 17]&[83.0, 45.0, 17.0, 17.0, 6.0]\\
CIAGAN&[100, 100, 100, 100, 83]&[100, 100, 100, 83, 67]&[100, 100, 100, 100, 67]&[100.0, 100.0, 100.0, 94.0, 72.0]\\
DP1&[0, 0, 0, 0, 0]&[67, 0, 0, 0, 0]&[100, 67, 67, 33, 0]&[56.0, 22.0, 22.0, 11.0, 0.0]\\
DP2&[83, 17, 0, 0, 0]&[83, 17, 0, 0, 0]&[100, 67, 67, 33, 0]&[89.0, 34.0, 22.0, 11.0, 0.0]\\
Fawkes-High&[100, 100, 100, 100, 83]&[100, 67, 67, 67, 67]&N/A&[100.0, 84.0, 84.0, 84.0, 75.0]\\
DP-Snow&[100, 83, 33, 0, 0]&[67, 67, 67, 0, 0]&N/A&[84.0, 75.0, 50.0, 0.0, 0.0]\\
Pixelation&[50, 0, 0, 0, 0]&[67, 0, 0, 0, 0]&N/A&[58.0, 0.0, 0.0, 0.0, 0.0]\\
K-Same-Pixel-ArcFace&[100, 100, 83, 67, 67]&[100, 100, 67, 67, 33]&N/A&[100.0, 100.0, 75.0, 67.0, 50.0]\\
mean&[73.0, 50.0, 40.0, 33.0, 29.0]&[92.0, 67.0, 54.0, 38.0, 21.0]&&\\
\hline
\textbf{gender}& BFW & DemogPairs&CelebA&mean \\
\hline
Original&[0, 0, 0, 0, 0]&[100, 0, 0, 0, 0]&[100, 0, 0, 0, 0]&[67.0, 0.0, 0.0, 0.0, 0.0]\\
CIAGAN&[100, 100, 100, 100, 100]&[100, 100, 100, 100, 100]&[100, 100, 100, 100, 100]&[100.0, 100.0, 100.0,100.0,100.0]\\
DP1&[100, 0, 0, 0, 0]&[100, 100, 0, 0, 0]&[100, 100, 100, 0, 0]&[100.0, 67.0, 33.0, 0.0, 0.0]\\
DP2&[100, 100, 0, 0, 0]&[100, 100, 0, 0, 0]&[100, 100, 100, 0, 0]&[100.0, 100.0, 33.0, 0.0, 0.0]\\
Fawkes-High&[100, 100, 100, 100, 100]&[100, 100, 100, 100, 100]&N/A&[100.0, 100.0, 100.0, 100.0, 100.0]\\
DP-Snow&[0, 0, 0, 0, 0]&[100, 100, 0, 0, 0]&N/A&[50.0, 50.0, 0.0, 0.0, 0.0]\\
Pixelation&[100, 0, 0, 0, 0]&[0, 0, 0, 0, 0]&N/A&[50.0, 0.0, 0.0, 0.0, 0.0]\\
K-Same-Pixel-ArcFace&[0, 0, 0, 0, 0]&[0, 0, 0, 0, 0]&N/A&[0, 0, 0, 0, 0]\\
mean&[62.0, 38.0, 25.0, 25.0, 25.0]&[75.0, 62.0, 25.0, 25.0, 25.0]&&\\

\hline
\end{tabular}
}
\end{table*}

\end{document}